%% file: main.tex
\documentclass[10pt]{article}

\PassOptionsToPackage{hyphens}{url}
\usepackage[preprint]{tmlr}

\usepackage[utf8]{inputenc}
\usepackage[T1]{fontenc}
\usepackage{microtype}
\usepackage{url}
\usepackage{hyperref}
\hypersetup{colorlinks=true, linkcolor=blue!50!black, citecolor=blue!50!black, urlcolor=blue!50!black}
\usepackage{booktabs}
\usepackage{amsmath}
\usepackage{amssymb}
\usepackage{graphicx}
\usepackage{enumitem}
\usepackage{multirow}
\usepackage{xcolor}

\input{results_macros}

\input{results_macros_verified}

\input{results_macros_mechanism}

\input{results_macros_localization}

\input{results_macros_tokens}
\input{results_macros_replay}
\input{results_macros_fitb}

\input{results_macros_expert}
\input{results_macros_truncation}

\input{results_macros_method}
\title{Same Simulation, Different Evidence: Where Retrieved Tool Output Costs LLM Agents Accuracy}

\author{\name Ke Zhang \email kzhan153@ucr.edu \\
      \addr University of California, Riverside
      \AND
      \name Sahchit Chundur \email smchundu@uci.edu \\
      \addr University of California, Irvine
      \AND
      \name Jerry Peprah Owusu \email jerrypeo@uci.edu \\
      \addr University of California, Irvine
      \AND
      \name Mohammad Javad Abdolhosseini Qomi \email mjaq@uci.edu \\
      \addr University of California, Irvine
      \AND
      \name Maziar Raissi \email maziarr@ucr.edu \\
      \addr University of California, Riverside}

\def\month{MM}  %
\def\year{YYYY} %
\def\openreview{\url{https://openreview.net/forum?id=XXXX}} %

\begin{document}

\maketitle

\begin{abstract}
Many LLM agent harnesses do not put a long tool output into the model's context window. They save it to a file, and the agent opens the file when it needs to. Earlier studies show that this form of context management can lower accuracy, but they do not explain why. We study the question with PHREEQC, a deterministic geochemistry simulator, on \VerN{} multiple-choice questions. We find one reason: the agent stops reading the simulator's report and answers from a file that it created itself. Both files contain the same result, but the agent's own file shows it with more digits. When the report is delivered as a file, GPT-5.1, GPT-5.2, and GPT-5.4 skip it 42--60\% of the time; Claude Sonnet 4.6 rarely does. We call this \emph{evidence substitution}. When these agents copy the longer number into their answer, they often cut off digits instead of rounding, and so choose a wrong option. Because the simulator is deterministic, we can re-run every input file that the agents wrote (\MechCallsN{} simulator calls) and locate the failure in each trajectory without a human or LLM judge. The agents compute the correct value about as often as before, but they answer with it less often. When we fix the simulation and change only how the report is delivered, GPT-5.2 loses \ReplayAbs{} points of accuracy. The fix is simple: tell the agent which file to trust. The agents then read the report again, and when they see the correct value they answer correctly 99\% of the time. Telling them to round correctly does not help. For anyone who builds agents that save tool output to files, this suggests a simple check: look at which file the agent actually reads before it answers, because the final accuracy alone does not show it.
\end{abstract}

\input{sections/introduction.tex}

\input{sections/setting.tex}
\input{sections/results_access.tex}
\input{sections/mechanism.tex}
\input{sections/hybrid.tex}
\input{sections/discussion.tex}
\input{sections/related_work.tex}
\input{sections/limitations.tex}
\input{sections/conclusion.tex}

\bibliography{references}
\bibliographystyle{tmlr}

\appendix

\input{appendix/item_verification.tex}
\input{appendix/conditions.tex}
\input{appendix/failure_localization.tex}
\input{appendix/contrasts.tex}
\input{appendix/frozen_replay.tex}
\input{appendix/fitb_evidence.tex}
\input{appendix/gemini_step_budget.tex}
\input{appendix/expert_review.tex}

\end{document}

%% file: results_macros.tex
\newcommand{\RawCapChars}{100{,}000}

\newcommand{\OutputMaxChars}{846.8k}

%% file: results_macros_verified.tex
\newcommand{\PoolN}{200}
\newcommand{\VerN}{158}

\newcommand{\VerMajority}{32.9\%}
\newcommand{\vOpusOneShot}{43.0\%}

\newcommand{\vOpusCoT}{39.9\%}

\newcommand{\vOpusTOC}{87.3\%}

\newcommand{\vOpusRaw}{88.0\%}

\newcommand{\vSonnetOneShot}{38.0\%}

\newcommand{\vSonnetCoT}{46.2\%}

\newcommand{\vSonnetTOC}{91.8\%}

\newcommand{\vSonnetRaw}{91.1\%}

\newcommand{\vGPTFiveFourOneShot}{25.9\%}

\newcommand{\vGPTFiveFourCoT}{34.2\%}

\newcommand{\vGPTFiveFourTOC}{75.3\%}

\newcommand{\vGPTFiveFourRaw}{70.9\%}

\newcommand{\vGPTFiveTwoOneShot}{27.2\%}

\newcommand{\vGPTFiveTwoCoT}{32.9\%}

\newcommand{\vGPTFiveTwoTOC}{57.0\%}

\newcommand{\vGPTFiveTwoRaw}{65.8\%}

\newcommand{\vGPTFiveOneOneShot}{41.1\%}

\newcommand{\vGPTFiveOneCoT}{36.1\%}

\newcommand{\vGPTFiveOneTOC}{57.0\%}

\newcommand{\vGPTFiveOneRaw}{67.7\%}

\newcommand{\vGeminiThreeFlashOneShot}{41.1\%}

\newcommand{\vGeminiThreeFlashCoT}{40.5\%}

\newcommand{\vGeminiThreeFlashTOC}{27.8\%}

\newcommand{\vGeminiThreeFlashRaw}{50.0\%}

\newcommand{\vOpusTOCMinusRaw}{$-$0.6\,pp}

\newcommand{\vSonnetTOCMinusRaw}{$+$0.6\,pp}

\newcommand{\vGPTFiveFourTOCMinusRaw}{$+$4.4\,pp}

\newcommand{\vGPTFiveTwoTOCMinusRaw}{$-$8.9\,pp}

\newcommand{\vGPTFiveOneTOCMinusRaw}{$-$10.8\,pp}

\newcommand{\vGeminiThreeFlashTOCMinusRaw}{$-$22.2\,pp}

\newcommand{\vToolGainMin}{15.8}
\newcommand{\vToolGainMax}{45.6}

\newcommand{\StrictN}{134}

\newcommand{\sOpusTOC}{86.6\%}
\newcommand{\sOpusRaw}{88.1\%}

\newcommand{\sSonnetTOC}{91.0\%}
\newcommand{\sSonnetRaw}{90.3\%}

\newcommand{\sGPTFiveFourTOC}{75.4\%}
\newcommand{\sGPTFiveFourRaw}{70.9\%}

\newcommand{\sGPTFiveTwoTOC}{59.7\%}
\newcommand{\sGPTFiveTwoRaw}{69.4\%}

\newcommand{\sGPTFiveOneTOC}{54.5\%}
\newcommand{\sGPTFiveOneRaw}{68.7\%}

\newcommand{\sGeminiThreeFlashTOC}{27.6\%}
\newcommand{\sGeminiThreeFlashRaw}{51.5\%}

\newcommand{\sOpusTOCMinusRaw}{$-$1.5\,pp}

\newcommand{\sSonnetTOCMinusRaw}{$+$0.7\,pp}

\newcommand{\sGPTFiveFourTOCMinusRaw}{$+$4.5\,pp}

\newcommand{\sGPTFiveTwoTOCMinusRaw}{$-$9.7\,pp}

\newcommand{\sGPTFiveOneTOCMinusRaw}{$-$14.2\,pp}

\newcommand{\fOpusTOCMinusRaw}{$+$0.5\,pp}

\newcommand{\fSonnetTOCMinusRaw}{$+$0.5\,pp}

\newcommand{\fGPTFiveFourTOCMinusRaw}{$+$2.5\,pp}

\newcommand{\fGPTFiveTwoTOCMinusRaw}{$-$7.5\,pp}

\newcommand{\fGPTFiveOneTOCMinusRaw}{$-$9.5\,pp}

\newcommand{\fOpusTOC}{83.5\%}
\newcommand{\fOpusRaw}{83.0\%}

\newcommand{\fSonnetTOC}{84.0\%}
\newcommand{\fSonnetRaw}{83.5\%}

\newcommand{\fGPTFiveFourTOC}{72.5\%}
\newcommand{\fGPTFiveFourRaw}{70.0\%}

\newcommand{\fGPTFiveTwoTOC}{54.0\%}
\newcommand{\fGPTFiveTwoRaw}{61.5\%}

\newcommand{\fGPTFiveOneTOC}{54.0\%}
\newcommand{\fGPTFiveOneRaw}{63.5\%}

\newcommand{\fGeminiThreeFlashTOC}{23.5\%}
\newcommand{\fGeminiThreeFlashRaw}{47.0\%}

\newcommand{\vSonnetTOCRunMin}{89.9\%}
\newcommand{\vSonnetTOCRunMax}{91.8\%}

\newcommand{\vGPTFiveFourTOCRunMin}{69.0\%}
\newcommand{\vGPTFiveFourTOCRunMax}{76.6\%}

\newcommand{\vGPTFiveTwoTOCRunMin}{55.7\%}
\newcommand{\vGPTFiveTwoTOCRunMax}{60.1\%}

\newcommand{\vGPTFiveOneTOCRunMin}{53.2\%}
\newcommand{\vGPTFiveOneTOCRunMax}{59.5\%}

\newcommand{\vSonnetGrep}{89.2\%}
\newcommand{\vGPTFiveTwoGrep}{57.6\%}
\newcommand{\vGPTFiveOneGrep}{48.1\%}
\newcommand{\vGPTFiveTwoRawRepeat}{67.7\%}

\newcommand{\vGPTFiveOneRawRepeat}{65.8\%}

%% file: results_macros_mechanism.tex
\newcommand{\hGPTFiveOneHybrid}{61.8\%}
\newcommand{\hGPTFiveOneHybridMin}{60.1\%}
\newcommand{\hGPTFiveOneHybridMax}{63.9\%}
\newcommand{\hGPTFiveTwoHybrid}{67.9\%}
\newcommand{\hGPTFiveTwoHybridMin}{66.5\%}
\newcommand{\hGPTFiveTwoHybridMax}{69.0\%}
\newcommand{\mGPTFiveOneRawAcc}{65.8\%}
\newcommand{\mGPTFiveOneTOCAcc}{55.5\%}
\newcommand{\mGPTFiveOneGrepAcc}{48.1\%}
\newcommand{\mGPTFiveTwoRawAcc}{67.7\%}
\newcommand{\mGPTFiveTwoTOCAcc}{58.0\%}
\newcommand{\mGPTFiveTwoGrepAcc}{57.6\%}

\newcommand{\declGPTFiveOneRaw}{1.9} \newcommand{\declGPTFiveOneTOC}{6.1}

\newcommand{\fitbDirect}{1.4\%} \newcommand{\fitbCoT}{2.8\%} \newcommand{\fitbTOC}{59.3\%}

\newcommand{\onceN}{93}
\newcommand{\onceCloTwo}{$-$9.8 [$-$15.4, $-$4.5]} \newcommand{\onceAccTwo}{$-$3.2 [$-$6.5, $-$0.0]}
\newcommand{\onceCloOne}{$-$6.8 [$-$11.2, $-$2.9]} \newcommand{\onceAccOne}{$-$4.9 [$-$8.0, $-$2.5]}
\newcommand{\onceCloFour}{$-$7.4 [$-$11.7, $-$3.8]} \newcommand{\onceAccFour}{$-$3.3 [$-$6.3, $-$0.2]}
\newcommand{\setupCIOne}{[$-$19.1, $-$2.9]} \newcommand{\setupCITwo}{[$-$16.2, $+$1.0]}
\newcommand{\looOne}{$-$14.0 to $-$9.7} \newcommand{\looTwo}{$-$10.9 to $-$5.8}

\newcommand{\ReplayAbs}{9.4}

%% file: results_macros_localization.tex
\newcommand{\MechTrajN}{5{,}773}
\newcommand{\MechCallsN}{11{,}007}
\newcommand{\MechInputsN}{10{,}605}
\newcommand{\MechValidOK}{4{,}971}
\newcommand{\MechValidN}{5{,}009}

\newcommand{\MechNumericN}{137}
\newcommand{\MechUnambigN}{115}
\newcommand{\MechBuildFailuresN}{1{,}039}
\newcommand{\MechBuildNoOptionN}{926}
\newcommand{\MechBuildDefaultRecoveredN}{5}
\newcommand{\MechReexecMovedN}{8}

\newcommand{\pSonnetTOC}{86.9} \newcommand{\seenSonnetTOC}{98.9} \newcommand{\cSonnetTOC}{98.0}
\newcommand{\pSonnetGrep}{86.1} \newcommand{\seenSonnetGrep}{98.3} \newcommand{\cSonnetGrep}{95.7}
\newcommand{\pGPTFiveFourRaw}{73.7} \newcommand{\seenGPTFiveFourRaw}{96.0} \newcommand{\cGPTFiveFourRaw}{99.0}
\newcommand{\pGPTFiveFourTOC}{80.0} \newcommand{\seenGPTFiveFourTOC}{91.2} \newcommand{\cGPTFiveFourTOC}{89.0}
\newcommand{\pGPTFiveFourRep}{78.5} \newcommand{\seenGPTFiveFourRep}{96.7} \newcommand{\cGPTFiveFourRep}{99.0}
\newcommand{\pGPTFiveTwoRaw}{65.0} \newcommand{\seenGPTFiveTwoRaw}{96.6} \newcommand{\cGPTFiveTwoRaw}{95.3}
\newcommand{\pGPTFiveTwoHybrid}{63.5} \newcommand{\seenGPTFiveTwoHybrid}{97.3} \newcommand{\cGPTFiveTwoHybrid}{94.9}
\newcommand{\pGPTFiveTwoTOC}{62.5} \newcommand{\seenGPTFiveTwoTOC}{90.7} \newcommand{\cGPTFiveTwoTOC}{82.4}
\newcommand{\pGPTFiveTwoTOCOct}{67.2} \newcommand{\seenGPTFiveTwoTOCOct}{92.4} \newcommand{\cGPTFiveTwoTOCOct}{83.5}
\newcommand{\pGPTFiveTwoRep}{63.1} \newcommand{\seenGPTFiveTwoRep}{95.4} \newcommand{\cGPTFiveTwoRep}{98.8}
\newcommand{\pGPTFiveTwoRound}{66.4} \newcommand{\seenGPTFiveTwoRound}{89.0} \newcommand{\cGPTFiveTwoRound}{81.5}
\newcommand{\pGPTFiveTwoGrep}{67.9} \newcommand{\seenGPTFiveTwoGrep}{86.0} \newcommand{\cGPTFiveTwoGrep}{82.5}
\newcommand{\pGPTFiveOneRaw}{52.6} \newcommand{\seenGPTFiveOneRaw}{98.6} \newcommand{\cGPTFiveOneRaw}{94.4}
\newcommand{\pGPTFiveOneHybrid}{49.6} \newcommand{\seenGPTFiveOneHybrid}{98.0} \newcommand{\cGPTFiveOneHybrid}{98.5}
\newcommand{\pGPTFiveOneTOC}{54.5} \newcommand{\seenGPTFiveOneTOC}{85.3} \newcommand{\cGPTFiveOneTOC}{85.3}
\newcommand{\pGPTFiveOneTOCOct}{50.4} \newcommand{\seenGPTFiveOneTOCOct}{81.2} \newcommand{\cGPTFiveOneTOCOct}{89.3}
\newcommand{\pGPTFiveOneRep}{50.7} \newcommand{\seenGPTFiveOneRep}{92.1} \newcommand{\cGPTFiveOneRep}{99.2}
\newcommand{\pGPTFiveOneRound}{56.9} \newcommand{\seenGPTFiveOneRound}{84.6} \newcommand{\cGPTFiveOneRound}{81.8}
\newcommand{\pGPTFiveOneGrep}{53.3} \newcommand{\seenGPTFiveOneGrep}{83.6} \newcommand{\cGPTFiveOneGrep}{78.7}

\newcommand{\shTotOneGrep}{$-$16.8 [$-$26.3, $-$8.0]} \newcommand{\shTotOneGrepP}{$-$16.8}
\newcommand{\shBldOneGrep}{$+$0.4 [$-$3.7, $+$4.9]} \newcommand{\shBldOneGrepP}{$+$0.4}
\newcommand{\shAccOneGrep}{$-$4.6 [$-$7.8, $-$1.9]} \newcommand{\shAccOneGrepP}{$-$4.6}
\newcommand{\shCloOneGrep}{$-$7.6 [$-$13.3, $-$2.2]} \newcommand{\shCloOneGrepP}{$-$7.6}
\newcommand{\shUnsOneGrep}{$-$5.0 [$-$11.3, $+$1.1]} \newcommand{\shUnsOneGrepP}{$-$5.0}
\newcommand{\shTotOneHyb}{$-$4.6 [$-$11.9, $+$2.4]} 
\newcommand{\shBldOneHyb}{$-$1.9 [$-$6.3, $+$3.0]} 
\newcommand{\shAccOneHyb}{$-$0.2 [$-$1.2, $+$0.9]} 
\newcommand{\shCloOneHyb}{$+$2.1 [$-$0.6, $+$5.1]} \newcommand{\shCloOneHybP}{$+$2.1}
\newcommand{\shUnsOneHyb}{$-$4.6 [$-$9.3, $+$0.1]} 
\newcommand{\shTotOneTOC}{$-$11.9 [$-$20.7, $-$2.7]} \newcommand{\shTotOneTOCP}{$-$11.9}
\newcommand{\shBldOneTOC}{$+$1.1 [$-$2.9, $+$5.6]} \newcommand{\shBldOneTOCP}{$+$1.1}
\newcommand{\shAccOneTOC}{$-$4.3 [$-$6.9, $-$2.2]} \newcommand{\shAccOneTOCP}{$-$4.3}
\newcommand{\shCloOneTOC}{$-$4.4 [$-$8.4, $-$0.5]} \newcommand{\shCloOneTOCP}{$-$4.4}
\newcommand{\shUnsOneTOC}{$-$4.2 [$-$11.0, $+$2.5]} \newcommand{\shUnsOneTOCP}{$-$4.2}
\newcommand{\shTotOneRep}{$-$3.6 [$-$11.7, $+$4.7]} 
\newcommand{\shBldOneRep}{$-$1.1 [$-$5.9, $+$3.4]} 
\newcommand{\shAccOneRep}{$-$2.2 [$-$4.3, $-$0.3]} 
\newcommand{\shCloOneRep}{$+$2.4 [$-$0.0, $+$5.2]} \newcommand{\shCloOneRepP}{$+$2.4}
\newcommand{\shUnsOneRep}{$-$2.7 [$-$9.3, $+$3.5]}

 \newcommand{\shCloOneRoundP}{$-$6.3}
 
\newcommand{\shTotOneTOCOct}{$-$12.4 [$-$22.6, $-$2.9]} 
\newcommand{\shBldOneTOCOct}{$-$1.2 [$-$6.1, $+$3.6]} 
\newcommand{\shAccOneTOCOct}{$-$5.5 [$-$9.2, $-$2.3]} 
\newcommand{\shCloOneTOCOct}{$-$2.4 [$-$6.7, $+$1.7]} \newcommand{\shCloOneTOCOctP}{$-$2.4}
\newcommand{\shUnsOneTOCOct}{$-$3.3 [$-$10.2, $+$3.2]} 
\newcommand{\shTotTwoGrep}{$-$11.7 [$-$21.2, $-$2.9]} \newcommand{\shTotTwoGrepP}{$-$11.7}
\newcommand{\shBldTwoGrep}{$+$1.8 [$-$2.7, $+$6.5]} \newcommand{\shBldTwoGrepP}{$+$1.8}
\newcommand{\shAccTwoGrep}{$-$4.7 [$-$8.7, $-$1.1]} \newcommand{\shAccTwoGrepP}{$-$4.7}
\newcommand{\shCloTwoGrep}{$-$7.8 [$-$13.6, $-$2.7]} \newcommand{\shCloTwoGrepP}{$-$7.8}
\newcommand{\shUnsTwoGrep}{$-$1.0 [$-$6.3, $+$4.0]} \newcommand{\shUnsTwoGrepP}{$-$1.0}
\newcommand{\shTotTwoHyb}{$-$1.0 [$-$8.0, $+$5.8]} 
\newcommand{\shBldTwoHyb}{$-$1.0 [$-$5.2, $+$2.9]} 
\newcommand{\shAccTwoHyb}{$+$0.3 [$-$1.2, $+$2.0]} 
\newcommand{\shCloTwoHyb}{$-$0.3 [$-$2.3, $+$2.0]} \newcommand{\shCloTwoHybP}{$-$0.3}
\newcommand{\shUnsTwoHyb}{$+$0.0 [$-$4.8, $+$4.4]} 
\newcommand{\shTotTwoTOC}{$-$11.4 [$-$21.2, $-$1.5]} \newcommand{\shTotTwoTOCP}{$-$11.4}
\newcommand{\shBldTwoTOC}{$-$1.5 [$-$5.6, $+$2.7]} \newcommand{\shBldTwoTOCP}{$-$1.5}
\newcommand{\shAccTwoTOC}{$-$2.5 [$-$5.3, $+$0.4]} \newcommand{\shAccTwoTOCP}{$-$2.5}
\newcommand{\shCloTwoTOC}{$-$7.7 [$-$12.6, $-$2.9]} \newcommand{\shCloTwoTOCP}{$-$7.7}
\newcommand{\shUnsTwoTOC}{$+$0.3 [$-$6.1, $+$6.1]} \newcommand{\shUnsTwoTOCP}{$+$0.3}
\newcommand{\shTotTwoRep}{$-$2.9 [$-$10.6, $+$5.5]} 
\newcommand{\shBldTwoRep}{$-$1.4 [$-$6.2, $+$4.3]} 
\newcommand{\shAccTwoRep}{$-$0.6 [$-$3.3, $+$1.6]} 
\newcommand{\shCloTwoRep}{$+$2.1 [$-$0.3, $+$5.0]} \newcommand{\shCloTwoRepP}{$+$2.1}
\newcommand{\shUnsTwoRep}{$-$3.1 [$-$8.0, $+$1.6]}

 \newcommand{\shCloTwoRoundP}{$-$8.5}
 
\newcommand{\shTotTwoTOCOct}{$-$5.1 [$-$16.1, $+$5.8]} 
\newcommand{\shBldTwoTOCOct}{$+$1.3 [$-$3.1, $+$6.0]} 
\newcommand{\shAccTwoTOCOct}{$-$1.7 [$-$4.7, $+$0.6]} 
\newcommand{\shCloTwoTOCOct}{$-$7.4 [$-$12.7, $-$2.2]} \newcommand{\shCloTwoTOCOctP}{$-$7.4}
\newcommand{\shUnsTwoTOCOct}{$+$2.7 [$-$3.4, $+$8.7]} 
\newcommand{\shTotFourTOC}{$-$4.4 [$-$12.4, $+$3.4]} \newcommand{\shTotFourTOCP}{$-$4.4}
\newcommand{\shBldFourTOC}{$+$4.1 [$+$0.0, $+$8.1]} \newcommand{\shBldFourTOCP}{$+$4.1}
\newcommand{\shAccFourTOC}{$-$2.6 [$-$4.8, $-$0.5]} \newcommand{\shAccFourTOCP}{$-$2.6}
\newcommand{\shCloFourTOC}{$-$7.2 [$-$11.1, $-$4.2]} \newcommand{\shCloFourTOCP}{$-$7.2}
\newcommand{\shUnsFourTOC}{$+$1.3 [$-$3.2, $+$5.8]} \newcommand{\shUnsFourTOCP}{$+$1.3}
\newcommand{\shTotFourRep}{$+$4.7 [$-$1.5, $+$10.9]} 
\newcommand{\shBldFourRep}{$+$3.4 [$-$0.8, $+$7.9]} \newcommand{\shBldFourRepP}{$+$3.4}
\newcommand{\shAccFourRep}{$+$0.4 [$-$1.0, $+$2.2]} 
\newcommand{\shCloFourRep}{$+$0.1 [$-$2.1, $+$2.2]} \newcommand{\shCloFourRepP}{$+$0.1}
\newcommand{\shUnsFourRep}{$+$0.9 [$-$3.4, $+$5.4]} 
\newcommand{\shCloTwoTOCUnamb}{$-$10.7}
\newcommand{\shAccTwoTOCUnamb}{$-$3.2}
\newcommand{\shCloOneTOCUnamb}{$-$5.8}
\newcommand{\shAccOneTOCUnamb}{$-$4.2}

\newcommand{\nsSonnetTOC}{10.5} \newcommand{\selSonnetTOC}{62.9}  \newcommand{\unrSonnetTOC}{8.5}
\newcommand{\nsSonnetGrep}{23.4} \newcommand{\selSonnetGrep}{60.1}  
\newcommand{\nsGPTFiveFourRaw}{0.0} \newcommand{\selGPTFiveFourRaw}{13.3} \newcommand{\defGPTFiveFourRaw}{77.8} \newcommand{\unrGPTFiveFourRaw}{3.1}
\newcommand{\nsGPTFiveFourTOC}{41.8} \newcommand{\selGPTFiveFourTOC}{61.0} \newcommand{\defGPTFiveFourTOC}{83.3} \newcommand{\unrGPTFiveFourTOC}{40.3}
\newcommand{\nsGPTFiveFourRep}{0.0} \newcommand{\selGPTFiveFourRep}{0.3} \newcommand{\defGPTFiveFourRep}{3.8} \newcommand{\unrGPTFiveFourRep}{0.0}
\newcommand{\nsGPTFiveTwoRaw}{0.0} \newcommand{\selGPTFiveTwoRaw}{29.7} \newcommand{\defGPTFiveTwoRaw}{100.0} \newcommand{\unrGPTFiveTwoRaw}{0.0}
\newcommand{\nsGPTFiveTwoTOC}{60.3} \newcommand{\selGPTFiveTwoTOC}{86.7}  \newcommand{\unrGPTFiveTwoTOC}{67.0}
\newcommand{\nsGPTFiveTwoGrep}{62.7} \newcommand{\selGPTFiveTwoGrep}{90.5}  
\newcommand{\nsGPTFiveTwoHybrid}{5.3} \newcommand{\selGPTFiveTwoHybrid}{34.8} \newcommand{\defGPTFiveTwoHybrid}{99.8} \newcommand{\unrGPTFiveTwoHybrid}{8.3}
\newcommand{\nsGPTFiveTwoTOCOct}{57.0} \newcommand{\selGPTFiveTwoTOCOct}{87.3} \newcommand{\defGPTFiveTwoTOCOct}{98.7} \newcommand{\unrGPTFiveTwoTOCOct}{63.5}
\newcommand{\nsGPTFiveTwoRep}{0.0} \newcommand{\selGPTFiveTwoRep}{1.3} \newcommand{\defGPTFiveTwoRep}{42.1} \newcommand{\unrGPTFiveTwoRep}{0.6}
\newcommand{\nsGPTFiveTwoRound}{55.1} \newcommand{\selGPTFiveTwoRound}{80.7} \newcommand{\defGPTFiveTwoRound}{98.4} \newcommand{\unrGPTFiveTwoRound}{59.9}
\newcommand{\nsGPTFiveOneRaw}{0.0} \newcommand{\selGPTFiveOneRaw}{29.7} \newcommand{\defGPTFiveOneRaw}{86.1} \newcommand{\unrGPTFiveOneRaw}{0.0}
\newcommand{\nsGPTFiveOneTOC}{53.0} \newcommand{\selGPTFiveOneTOC}{67.5}  \newcommand{\unrGPTFiveOneTOC}{47.6}
\newcommand{\nsGPTFiveOneGrep}{41.8} \newcommand{\selGPTFiveOneGrep}{53.8}  
\newcommand{\nsGPTFiveOneHybrid}{1.1} \newcommand{\selGPTFiveOneHybrid}{23.0} \newcommand{\defGPTFiveOneHybrid}{74.3} \newcommand{\unrGPTFiveOneHybrid}{1.5}
\newcommand{\nsGPTFiveOneTOCOct}{48.7} \newcommand{\selGPTFiveOneTOCOct}{63.3} \newcommand{\defGPTFiveOneTOCOct}{81.0} \newcommand{\unrGPTFiveOneTOCOct}{39.3}
\newcommand{\nsGPTFiveOneRep}{1.6} \newcommand{\selGPTFiveOneRep}{10.4} \newcommand{\defGPTFiveOneRep}{38.0} \newcommand{\unrGPTFiveOneRep}{0.8}
\newcommand{\nsGPTFiveOneRound}{45.9} \newcommand{\selGPTFiveOneRound}{64.6} \newcommand{\defGPTFiveOneRound}{81.0} \newcommand{\unrGPTFiveOneRound}{46.2}

\newcommand{\trapSonnetTOC}{1/35}  
\newcommand{\trapSonnetGrep}{2/15}  
\newcommand{\trapGPTFiveFourRaw}{0/0}  
\newcommand{\trapGPTFiveFourTOC}{15/31}  
\newcommand{\trapGPTFiveFourRep}{0/0}  
\newcommand{\trapGPTFiveTwoRaw}{1/2}  
\newcommand{\trapGPTFiveTwoTOC}{26/35}  
\newcommand{\trapGPTFiveTwoGrep}{7/13}  
\newcommand{\trapGPTFiveTwoHybrid}{6/12}  
\newcommand{\trapGPTFiveTwoTOCOct}{11/15}  
\newcommand{\trapGPTFiveTwoRep}{0/0}  
\newcommand{\trapGPTFiveTwoRound}{17/25}  
\newcommand{\trapGPTFiveOneRaw}{2/3}  
\newcommand{\trapGPTFiveOneTOC}{19/27}  
\newcommand{\trapGPTFiveOneGrep}{5/6}  
\newcommand{\trapGPTFiveOneHybrid}{1/1}  
\newcommand{\trapGPTFiveOneTOCOct}{3/5}  
\newcommand{\trapGPTFiveOneRep}{0/0}  
\newcommand{\trapGPTFiveOneRound}{14/19}  

\newcommand{\expoSkipGPTFiveTwo}{6 of 7}
\newcommand{\expoSkipGPTFiveOne}{14 of 17}

\newcommand{\medExSonnet}{99.7}  \newcommand{\medUnSonnet}{83.3}   
\newcommand{\medExGPTFiveFour}{99.4}  \newcommand{\medUnGPTFiveFour}{75.2}   
\newcommand{\medExGPTFiveTwo}{100.0}  \newcommand{\medUnGPTFiveTwo}{72.5}   
\newcommand{\medExGPTFiveOne}{100.0}  \newcommand{\medUnGPTFiveOne}{66.3}   

\newcommand{\medExGPTFiveFourRaw}{98.9}
\newcommand{\medUnGPTFiveFourRaw}{100.0}

\newcommand{\chanRepGPTFiveFourTOC}{63.9}
\newcommand{\chanRepGPTFiveFourRep}{100.0}
\newcommand{\chanRepGPTFiveTwoRaw}{98.8}

\newcommand{\chanRepGPTFiveTwoTOCOct}{40.8}
\newcommand{\chanRepGPTFiveTwoRep}{99.4}
\newcommand{\chanRepGPTFiveTwoHybrid}{92.5}
\newcommand{\chanRepGPTFiveOneRaw}{100.0}

\newcommand{\chanRepGPTFiveOneTOCOct}{68.0}
\newcommand{\chanRepGPTFiveOneRep}{99.2}
\newcommand{\chanRepGPTFiveOneHybrid}{99.0}

\newcommand{\effStepsRaw}{7.2} 
\newcommand{\effStepsTOC}{9.1}

\newcommand{\accGPTFiveOneTOCOct}{53.8\%}
\newcommand{\accGPTFiveOneRep}{61.7\%}
\newcommand{\accGPTFiveOneRepRuns}{62.7\% and 60.8\%}
\newcommand{\accGPTFiveTwoTOCOct}{63.3\%}
\newcommand{\accGPTFiveTwoRep}{66.8\%}
\newcommand{\accGPTFiveTwoRepRuns}{67.1\% and 66.5\%}

\newcommand{\dRepOctGPTFiveOneP}{$+$7.9}

\newcommand{\dRepTOCGPTFiveOneP}{$+$6.3}
\newcommand{\dRepRawGPTFiveOne}{$-$5.1\,pp [$-$11.4, $+$0.9]}

\newcommand{\dHybOctGPTFiveOneP}{$+$8.0}

\newcommand{\dHybRawGPTFiveOneP}{$-$5.0}

\newcommand{\dRepOctGPTFiveTwoP}{$+$3.5}

\newcommand{\dRepTOCGPTFiveTwoP}{$+$7.9}

\newcommand{\dHybOctGPTFiveTwoP}{$+$4.6}

\newcommand{\dHybRawGPTFiveTwoP}{$+$1.2}

\newcommand{\accGPTFiveFourRep}{81.6\%}
\newcommand{\accGPTFiveFourRepRuns}{81.6\% and 81.6\%}
\newcommand{\mGPTFiveFourTOCAcc}{73.4\%}
\newcommand{\accGPTFiveFourRawLogged}{76.6\%}
\newcommand{\dRepTOCGPTFiveFour}{$+$7.8\,pp [$+$2.2, $+$13.6]}
\newcommand{\dRepTOCGPTFiveFourP}{$+$7.8}
\newcommand{\dRepRawGPTFiveFour}{$+$7.9\,pp [$+$2.5, $+$13.3]}

\newcommand{\accGPTFiveOneRound}{56.6\%}
\newcommand{\accGPTFiveOneRoundRuns}{55.7\% and 57.6\%}

\newcommand{\dRoundOctGPTFiveOneP}{$+$2.8}

\newcommand{\dRoundTOCGPTFiveOneP}{$+$1.2}
\newcommand{\dRepRoundGPTFiveOne}{$+$5.1\,pp [$-$1.6, $+$12.0]}

\newcommand{\accGPTFiveTwoRound}{59.5\%}
\newcommand{\accGPTFiveTwoRoundRuns}{62.0\% and 57.0\%}

\newcommand{\dRoundOctGPTFiveTwoP}{$-$3.8}

\newcommand{\dRoundTOCGPTFiveTwoP}{$+$0.6}
\newcommand{\dRepRoundGPTFiveTwo}{$+$7.3\,pp [$+$0.6, $+$13.9]}

\newcommand{\dRepTOCBoth}{$+$7.1\,pp [$+$1.9, $+$12.4]}

\newcommand{\dTOCRawOpus}{$-$0.6\,pp [$-$3.8, $+$2.5]}
\newcommand{\dTOCRawOpusP}{$-$0.6}
\newcommand{\dTOCRawSonnet}{$-$0.5\,pp [$-$3.8, $+$3.2]}
\newcommand{\dTOCRawSonnetP}{$-$0.5}
\newcommand{\dTOCRawGPTFiveFour}{$+$0.2\,pp [$-$5.5, $+$5.7]}
\newcommand{\dTOCRawGPTFiveFourP}{$+$0.2}
\newcommand{\dTOCRawGPTFiveTwo}{$-$7.9\,pp [$-$14.9, $-$0.9]}
\newcommand{\dTOCRawGPTFiveTwoP}{$-$7.9}
\newcommand{\dTOCRawGPTFiveOne}{$-$11.3\,pp [$-$18.0, $-$4.9]}
\newcommand{\dTOCRawGPTFiveOneP}{$-$11.3}

\newcommand{\pHolmTOCRawOpus}{1.000}

\newcommand{\pHolmTOCRawSonnet}{1.000}

\newcommand{\pHolmTOCRawGPTFiveFour}{1.000}
\newcommand{\pTOCRawGPTFiveTwo}{0.032}
\newcommand{\pHolmTOCRawGPTFiveTwo}{0.128}

\newcommand{\pHolmTOCRawGPTFiveOne}{0.005}

%% file: results_macros_tokens.tex
\newcommand{\tokOpusTOC}{--}
\newcommand{\tokSonnetTOC}{100.1k}
\newcommand{\tokGPTFiveFourTOC}{55.7k}
\newcommand{\tokGPTFiveFourRaw}{64.3k}
\newcommand{\tokGPTFiveTwoTOC}{61.8k} \newcommand{\tokGPTFiveTwoRaw}{65.7k} \newcommand{\tokGPTFiveTwoHybrid}{59.7k}
\newcommand{\tokGPTFiveOneTOC}{52.8k} \newcommand{\tokGPTFiveOneRaw}{60.9k} \newcommand{\tokGPTFiveOneHybrid}{28.3k$^\dagger$}
\newcommand{\tokHybridText}{For GPT-5.2, Hybrid used 59.7k input tokens per item, against 61.8k under TOC and 65.7k inline, with the saving relative to inline coming from long outputs. GPT-5.1's Hybrid runs used far fewer tokens (28.3k per item), mostly because the agent rarely read an optional database reference file that it read in about two thirds of its other TOC and inline runs; the TOC control run also used fewer tokens (34.2k), so we do not claim this as a saving of Hybrid.}
\newcommand{\tokGPTFiveTwoTOCOct}{58.0k} \newcommand{\tokGPTFiveTwoRep}{49.7k}
\newcommand{\tokGPTFiveOneTOCOct}{34.2k} \newcommand{\tokGPTFiveOneRep}{30.8k}

%% file: results_macros_replay.tex
\newcommand{\ReplayItems}{131}
\newcommand{\ReplayRuns}{3}
\newcommand{\ReplayBranches}{786}
\newcommand{\ReplayPairs}{393}
\newcommand{\ReplaySetups}{19}
\newcommand{\ReplayFullCorrect}{280}
\newcommand{\ReplayTOCCorrect}{243}

\newcommand{\ReplayFullAcc}{71.2}
\newcommand{\ReplayTOCAcc}{61.8}
\newcommand{\ReplayDelta}{$-$9.4}
\newcommand{\ReplayCI}{[$-$14.5, $-$4.6]}
\newcommand{\ReplayClusterCI}{[$-$13.9, $-$5.2]}

\newcommand{\ReplayFitN}{115}

\newcommand{\ReplayFitFullAcc}{75.4}
\newcommand{\ReplayFitTOCAcc}{65.2}
\newcommand{\ReplayFitDelta}{$-$10.1}
\newcommand{\ReplayLongN}{16}

\newcommand{\ReplayLongFullAcc}{41.7}
\newcommand{\ReplayLongTOCAcc}{37.5}

\newcommand{\ReplayKeyN}{84}

\newcommand{\ReplayKeyFullAcc}{93.7}
\newcommand{\ReplayKeyTOCAcc}{83.7}

\newcommand{\ReplayRunOneFullCorrect}{92}
\newcommand{\ReplayRunOneTOCCorrect}{81}

\newcommand{\ReplayRunOneDelta}{$-$8.4}
\newcommand{\ReplayRunTwoFullCorrect}{93}
\newcommand{\ReplayRunTwoTOCCorrect}{79}

\newcommand{\ReplayRunTwoDelta}{$-$10.7}
\newcommand{\ReplayRunThreeFullCorrect}{95}
\newcommand{\ReplayRunThreeTOCCorrect}{83}

\newcommand{\ReplayRunThreeDelta}{$-$9.2}

%% file: results_macros_fitb.tex
\newcommand{\fitbVerN}{145}
\newcommand{\fitbSkipN}{97} \newcommand{\fitbSkipPct}{66.9} \newcommand{\fitbReadN}{48}
\newcommand{\fitbOwnSkip}{97 of 97} \newcommand{\fitbOwnRead}{14 of 48}
\newcommand{\fitbMoreSkip}{87 of 97} \newcommand{\fitbMoreRead}{7 of 48}
\newcommand{\fitbCorrSkip}{55 of 97} \newcommand{\fitbCorrRead}{31 of 48}

%% file: results_macros_expert.tex
\newcommand{\humanN}{75} \newcommand{\humanSetups}{15}
\newcommand{\humanCorrect}{60} \newcommand{\humanAcc}{80.0\%}
\newcommand{\humanAbstain}{13} \newcommand{\humanWrong}{2} 
\newcommand{\expertErrN}{55} \newcommand{\expertErrVerN}{39}
\newcommand{\expertPresentN}{12} \newcommand{\expertPresentPct}{22\%}

%% file: results_macros_truncation.tex
\newcommand{\truncFellN}{60}  %
\newcommand{\truncLit}{105 of 145}
\newcommand{\truncLitPct}{72}

\newcommand{\truncConv}{12 of 31}
\newcommand{\truncConvPct}{39}

\newcommand{\truncGradedOne}{$-$10.8 [$-$18.5, $-$3.3]}

\newcommand{\truncForgivenOne}{$-$8.5 [$-$16.0, $-$1.1]}
\newcommand{\truncForgivenOneP}{$-$8.5}
\newcommand{\truncGradedTwo}{$-$8.4 [$-$16.9, $+$0.0]}

\newcommand{\truncForgivenTwo}{$-$2.8 [$-$10.8, $+$4.9]}

\newcommand{\truncGradedFour}{$-$3.2 [$-$10.5, $+$3.6]}

\newcommand{\truncForgivenFour}{$+$0.4 [$-$6.3, $+$7.2]}

\newcommand{\bothTrapGPT}{3 of 12}  %
\newcommand{\bothTrapSonnet}{32 of 33}  %

%% file: sections/introduction.tex
\section{Introduction}
\label{sec:intro}

\begin{figure}[t]
\centering
\includegraphics[width=\linewidth]{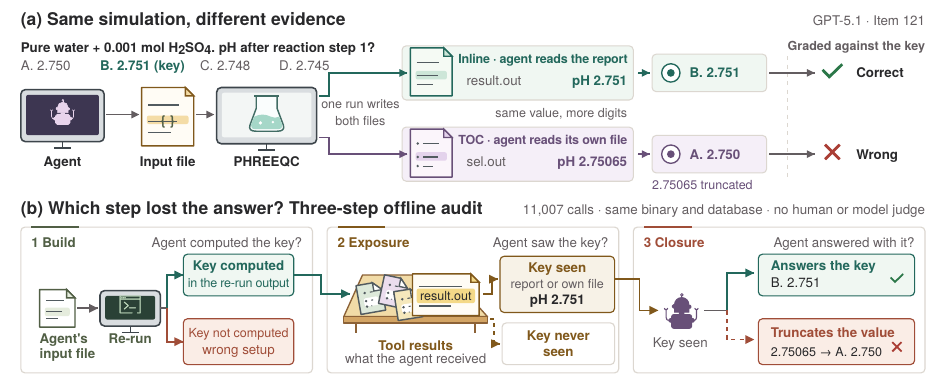}
\caption{One simulation writes the correct value twice, and which copy the agent reads decides the answer. (a) Item 121, GPT-5.1. The same run prints the key in the simulator's report (\texttt{result.out}, pH 2.751) and in the output file the agent defined itself (\texttt{sel.out}, 2.75065). With the report inline the agent reads it and answers B. Under TOC it does not open the report, reads its own file, truncates the value, and answers A. (b) The offline audit that assigns each wrong answer to a step. We re-run every input the agent wrote (\MechCallsN{} calls, same binary and database) and ask three questions. Did the agent compute the key, that is, does a re-run print it (build)? Did the agent see the key in a tool result it received, from the report or from its own file (exposure)? Did the agent answer with it (closure)? In (a) the TOC trajectory passes the first two and fails the third. No human or model judge is involved.}
\label{fig:method}
\label{fig:evidence}
\end{figure}

LLM agents increasingly call tools whose outputs are long, such as logs, search results, and simulator reports. A harness must choose how such output reaches the model: paste it into the context, truncated to a fixed budget (\emph{inline}), or save it to a file that the agent reads through an outline, a search tool, or ranged reads (\emph{retrieval}). Production harnesses now write long tool results to files and return a pointer or a short preview, and report the tokens that this saves \citep{anthropic2025context,katz2026dynamic,curme2026deepagents}. The effect on accuracy is less clear. Controlled comparisons find that file-based delivery lowers accuracy for some models and harnesses and not for others \citep{sen2026grep,mcmillan2026structured}, and on the items used here, \citet{zhang2026phreeqcmcq} report that a table-of-contents interface lowers the accuracy of GPT-5.1 and GPT-5.2 but not of GPT-5.4 or the Claude models. These studies report final accuracy and offer explanations that they do not test, such as an added step of locating and integrating the file or a limited ability to navigate structured output. What remains unknown is where the loss arises: a wrong answer rarely shows whether the agent computed the wrong value, never saw the right one, or saw it and still answered wrongly.

A deterministic simulator makes this observable. We study LLM agents operating PHREEQC, an aqueous-geochemistry simulator \citep{parkhurst2013phreeqc} whose section-structured reports reach \OutputMaxChars{} characters, on \VerN{} multiple-choice items whose answer keys we re-verified by execution (Section~\ref{sec:items}). We change only how the simulator's report reaches the agent. Our main comparison has two conditions: the report is pasted into the tool result (inline), or the agent receives a table of contents (TOC) of the report's sections and reads the lines it wants. Section~\ref{sec:harness} lists all six conditions, which add keyword search and three variants of the TOC condition. Each simulation writes its result to a report. If the agent asks for it, the simulation also writes the same result to a second file, with more digits. We call the value of the correct option the \emph{keyed value}; both files contain it. Under TOC, the GPT models answer from the second file and not from the report (Figure~\ref{fig:method}).

Because the simulator is deterministic, we can re-run every input an agent wrote. We also search every tool result that the agent received. This gives each trajectory on a numeric item three labels, with no human or model judge: did the agent compute the keyed value (\emph{build}), did it see that value (\emph{exposure}), and did it answer with it (\emph{closure})? Each stage needs a different fix. A build loss needs better simulation inputs. An exposure or closure loss needs a change in what the agent reads. We therefore first locate the loss and then test the location by intervening on that stage. Our contributions are as follows.
\begin{enumerate}[leftmargin=*, itemsep=2pt]
    \item \textbf{Evidence substitution.} Under TOC, GPT-5.1, GPT-5.2, and GPT-5.4 leave the simulator's report unopened in 42--60\% of trajectories and answer from their own \texttt{SELECTED\_OUTPUT} files, which hold the same value with more digits. When that value falls between two options, they often choose the truncated one. The loss was earlier attributed to an inability to navigate the report \citep{zhang2026phreeqcmcq}; we find that the report is often not opened (Section~\ref{sec:mechanism}).
    \item \textbf{An audit that locates the loss without a judge.} We re-run all \MechCallsN{} simulator calls that the agents made. For each trajectory on a numeric item, this shows what the agent computed, what it saw, and what it answered. The result is simple: the agents compute the correct value about as often with TOC as with inline output, but they answer with it less often. When we continue the same simulation under both conditions, TOC still costs GPT-5.2 \ReplayAbs\,pp (Sections~\ref{sec:localize} and~\ref{sec:mechanism}).
    \item \textbf{An intervention that tests the location.} An instruction that names the report as authoritative removes report skipping and restores closure to 99\% without changing build, and GPT-5.2 returns to its inline accuracy. Inlining reports that fit has the same effect; an instruction that only asks for correct rounding has none (Section~\ref{sec:hybrid}).
\end{enumerate}

%% file: sections/setting.tex
\section{Setting}
\label{sec:benchmark}

We use PHREEQC because three of its properties let us observe each step of a tool call. First, it is deterministic: re-running an input file that an agent wrote reproduces exactly what the agent's simulation printed. Second, its report is long and divided into sections, so retrieval is a natural way to deliver it. Third, an agent can ask it to write chosen quantities to a second file, so the same value can exist in two files in two different forms.

\subsection{Items and Verification}
\label{sec:items}

We use the four-option multiple-choice items of \citet{zhang2026phreeqcmcq}, written by one domain expert. Each describes a PHREEQC simulation setup and asks for one quantity from its output (a pH at a reaction step, a saturation index, a molality in a transport cell); several items share a setup. Reference answers come from PHREEQC 3.8.6 with a fixed, author-supplied \texttt{phreeqc.dat}. For numeric items, distractors typically differ from the key in the last printed digit, so they can be answered only by computing the value, not by estimating it; the remaining items ask for a comparison, a name, or a count.

\paragraph{Answer-key verification.}
We re-ran all \PoolN{} items of the original pool in PHREEQC and keep the \VerN{} items whose printed value matches the keyed option exactly and uniquely. All results in the main text use these \VerN{} items. Appendix~\ref{sec:appendix-verification} gives the procedure and shows that the full pool leads to the same conclusions. As a reference point, a PHREEQC user with eight years of experience who did not write the items, working with the simulator, answered \humanCorrect{} of \humanN{} sampled verified items correctly and abstained on \humanAbstain{} whose simulation matched no option (Appendix~\ref{sec:appendix-expert}).

\subsection{Agents and Output Delivery}
\label{sec:harness}

Each agent has four tools --- \texttt{write\_file}, \texttt{read\_file} (optionally with a line range), \texttt{list\_file}, and \texttt{execute\_phreeqc} --- and works in an isolated workspace for at most 24 steps, committing to an answer by writing it to a file. PHREEQC writes a formatted report (\texttt{result.out}) that prints each quantity at the precision the answer options use; agents may also add \texttt{SELECTED\_OUTPUT} or \texttt{USER\_PUNCH} blocks, which write chosen quantities to separate files with more digits. The conditions differ in what \texttt{execute\_phreeqc} returns and in the system-prompt line that describes it (Table~\ref{tab:delivery}; Appendix~\ref{sec:appendix-intervention}). We refer to the first as \emph{inline} and to TOC and Grep as \emph{retrieval}.

\begin{table}[t]
\centering
\footnotesize
\setlength{\tabcolsep}{3pt}
\begin{tabular}{p{0.24\linewidth}p{0.70\linewidth}}
\toprule
Condition & What \texttt{execute\_phreeqc} returns \\
\midrule
Inline & The report, head/tail-truncated to \RawCapChars{} characters \\
TOC & Metadata and an index of report sections with starting lines; the agent reads line ranges \\
Grep & Metadata only; the agent searches the report with regular expressions \\
Hybrid & The report when it fits within \RawCapChars{} characters, the TOC otherwise \\
TOC+Report & TOC, with an instruction to answer from the report (Section~\ref{sec:hybrid}) \\
TOC+Round & TOC, with an instruction to round to the options' precision (Section~\ref{sec:hybrid}) \\
\bottomrule
\end{tabular}
\caption{The six ways in which the simulator's report is delivered. The simulation, the tools, and the step budget are the same in all of them.}
\label{tab:delivery}
\end{table}

We evaluate Claude Opus 4.6, Claude Sonnet 4.6, GPT-5.4, GPT-5.2, GPT-5.1, and Gemini 3 Flash Preview through their APIs: the Claude models at temperature 0 without extended thinking, the GPT and Gemini models at API defaults with no reasoning-effort setting. Without the simulator, no model exceeds \vSonnetCoT{} with direct or chain-of-thought prompting (the majority-label rate is \VerMajority), so the items require the tool (Appendix~\ref{sec:appendix-intervention}, Table~\ref{tab:main-results}). Gemini 3 Flash gives no parseable answer on 46\% of items even inline, so we report it in Appendix~\ref{sec:appendix-gemini-step-budget} and leave it out of the analyses below.

\paragraph{Runs.}
Each model has one headline inline run and one headline TOC run. We rerun TOC three times for Sonnet and the GPT models and inline once for the GPT models, and run Grep and the interventions on a subset of models (Table~\ref{tab:runs}). Rates are averaged over runs; counts use each trajectory once. Our primary comparison is TOC against inline per model, with every run pooled per item, 95\% item-bootstrap intervals, and sign-flip permutation tests Holm-corrected over the five models.

\subsection{Locating Where an Agent Fails}
\label{sec:localize}

For every agent run we store the full trace: the model's messages, every tool call, and every tool result. We analyze these logs offline. For every full-agent trajectory with a complete log (\MechTrajN) we rebuilt each input file as it was when the agent executed it and re-ran it with the same binary and database: \MechCallsN{} simulator calls in all. As a check, we compared the re-run of the last execution with the stored report in \MechValidN{} numeric trajectories. The two are identical in \MechValidOK; the rest differ only in line order. For the \MechNumericN{} items with numeric answers we then ask three questions (Figure~\ref{fig:method}), matching values at the precision the options are printed with:
\begin{itemize}[leftmargin=*, itemsep=1pt]
    \item \textbf{Build} ($p$): did the agent compute the keyed value, that is, did any of its own simulations print it?
    \item \textbf{Exposure} ($s$): given that it was computed, did the agent see the value, inline or through a read or search result?
    \item \textbf{Closure} ($c$): given that the agent saw the value, did it answer with it, that is, choose the keyed option?
\end{itemize}
Accuracy equals $p\,s\,c$ plus the share of answers that were correct without the keyed value in view. We call these answers \emph{unsupported}; they are nearest-option picks and guesses. Appendix~\ref{sec:appendix-contrasts} attributes each condition's accuracy change to these four terms. We also record the file from which each seen value came. Matching is numeric, so a number can also match on a line that prints another quantity. Two checks address this. Automatic and manual stage labels agree on 52 of 58 numeric trajectories that we read by hand. And the results hold on the \onceN{} numeric items whose keyed value occurs only once in the reference report (Appendix~\ref{sec:appendix-localization}).

%% file: sections/results_access.tex
\section{The Cost of Retrieval Depends on the Model}
\label{sec:main-results}

\begin{figure}[t]
\centering
\includegraphics[width=0.75\linewidth]{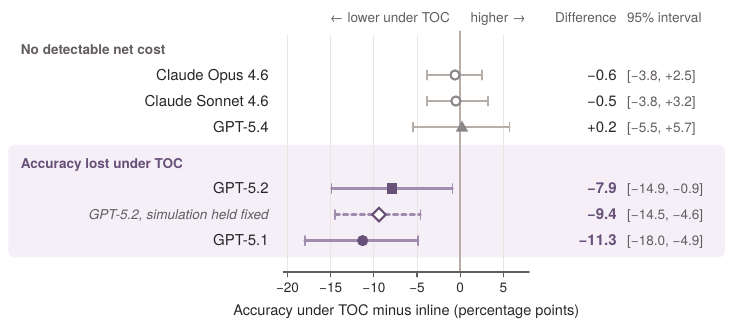}
\caption{Net cost of TOC delivery: accuracy under TOC minus inline on the \VerN{} items, with every run pooled per item. Bars are 95\% item-bootstrap intervals; the two columns on the right give the difference and its interval (also in Table~\ref{tab:main-results}). The shaded group contains the rows whose interval excludes zero (Holm-corrected tests in Table~\ref{tab:main-results}). The dashed row is the paired replay of GPT-5.2 (\ReplayItems{} numeric items), in which both arms continue from the same executed simulation, so only the delivery differs (Section~\ref{sec:mechanism}).}
\label{fig:net-cost}
\end{figure}

\paragraph{\texorpdfstring{GPT-5.1 and GPT-5.2 lose accuracy under TOC.}{GPT-5.1 and GPT-5.2 lose accuracy under TOC.}}
Pooling every run per item, TOC changes accuracy relative to inline output by \dTOCRawGPTFiveOneP\,pp for GPT-5.1 and \dTOCRawGPTFiveTwoP\,pp for GPT-5.2 (Figure~\ref{fig:net-cost}). \citet{zhang2026phreeqcmcq} reported the same net effect. We reproduce it on verified items and with reruns. The GPT-5.1 difference is significant after Holm correction over the five models. The GPT-5.2 difference is not, but with the simulation held fixed the change is \ReplayDelta\,pp, with an interval that excludes zero (Section~\ref{sec:mechanism}). One keyword-search run, which uses no PHREEQC section structure, shows a similar cost (Table~\ref{tab:master}).

\paragraph{Claude Opus 4.6, Claude Sonnet 4.6, and GPT-5.4 show no detectable net cost.}
Their pooled differences are within 1\,pp of zero, with intervals that include zero (Figure~\ref{fig:net-cost}).

%% file: sections/mechanism.tex
\section{Agents Answer from Their Own Output}
\label{sec:mechanism}
\label{sec:diagnostic}

\input{sections/table_master.tex}

\paragraph{Under retrieval, agents substitute their own output for the report.}
Under TOC, GPT-5.2 and GPT-5.1 \emph{skip} the report in more than half of their trajectories: no part of it reaches them. In most trajectories they read their own selected-output files (Table~\ref{tab:master}, Skip and Own; Figure~\ref{fig:method}a). They define such files about as often inline (Table~\ref{tab:intervention}, Def.), but there the report is already in context and they answer from it. We call this \emph{evidence substitution}: leaving the tool's canonical report unopened and answering from output that the agent defined itself. Sonnet skips the report in about one of ten TOC trajectories. Figure~\ref{fig:trajectory} shows one trajectory of this kind in full: the agent defines a selected-output file in its first input, reads that file after the run, and answers with the truncation of the value it prints, although the index it received points to the section of the report that prints the keyed value.

\begin{figure}[t]
\centering
\footnotesize
\begin{tabular}{@{}r l@{}}
\toprule
Step & Agent call $\rightarrow$ tool result (abridged) \\
\midrule
1 & \parbox[t]{0.92\linewidth}{\raggedright \texttt{write\_file(calc.inp)}: \texttt{SOLUTION 1 Pure water / pH 7 charge / REACTION 1 / H2SO4 1 / 0.001 / SELECTED\_OUTPUT -file sel.out -pH true -solution true / END}} \\
2 & \parbox[t]{0.92\linewidth}{\raggedright \texttt{execute\_phreeqc(calc.inp)} $\rightarrow$ \texttt{ok; output\_files: phreeqc.log, result.out (6,957 bytes), sel.out (315 bytes); result\_out\_toc: 19 entries, among them \{line 92: Reaction step 1\}, \{line 112: Solution composition\}, \{line 118: Description of solution\}; hint: Use read\_file with start\_line/end\_line to read specific sections.}} \\
3 & \parbox[t]{0.92\linewidth}{\raggedright \texttt{read\_file(sel.out, 1, 200)} $\rightarrow$ \texttt{sim state soln \ldots{} step pH pe / 1 i\_soln 1 \ldots{} 6.99738 4 / 1 react 1 \ldots{} 1 2.75065 14.6924}} \\
4 & \parbox[t]{0.92\linewidth}{\raggedright \texttt{write\_file(final\_answer.txt)}: \texttt{\textless{}\textless{}\textless{} A \textgreater{}\textgreater{}\textgreater{}}} \\
\midrule
-- & \parbox[t]{0.92\linewidth}{\raggedright Not read: \texttt{result.out}, line 120 (in ``Description of solution'' of reaction step 1): \texttt{pH = 2.751}} \\
\bottomrule
\end{tabular}
\caption{A GPT-5.1 TOC trajectory on item 121 (pH after the first reaction step; options A 2.750, B 2.751, C 2.748, D 2.745; key B), with tool results abridged and ``/'' marking line breaks. The agent answers A, the truncation of the 2.75065 in its own selected-output file, and never reads the report, which prints the keyed 2.751.}
\label{fig:trajectory}
\end{figure}

\paragraph{Answering from one's own output fails in two ways.}
\emph{Closure.} The report prints each quantity at the precision the options use, whereas selected-output files print more digits, so an agent reading them must round, and the GPT models often truncate instead. A typical case reads ``gives pH = 2.75065, which rounds to 2.750'' for a keyed value of 2.751. We call a trajectory a \emph{trap case} when the value the agent saw rounds to the keyed value but truncates to another option. In trap cases under TOC, GPT-5.2 chose the truncated option \trapGPTFiveTwoTOC{} times, GPT-5.1 \trapGPTFiveOneTOC, and Sonnet \trapSonnetTOC{} (Table~\ref{tab:master}, Trap). For the GPT models the truncated value always came from the agent's own file, never from the report. They chose the truncated option more often when it is a literal prefix of the printed number (\truncLitPct\%) than when the two are printed in different notations (\truncConvPct\%). This suggests digit matching, not rounding (Appendix~\ref{sec:appendix-contrasts}). Counting every truncated answer as correct removes most of the TOC-minus-inline difference of GPT-5.2 and about a fifth of GPT-5.1's (Table~\ref{tab:truncation}). Section~\ref{sec:hybrid} tests whether this is only a rounding error.
\emph{Exposure.} The remaining loss arises one step earlier. Sometimes the agent computed the keyed value but never saw it. In most of these cases it had skipped the report, and its own file printed a different quantity, such as a log activity, from which it answered (Appendix~\ref{sec:appendix-contrasts}).

\paragraph{\texorpdfstring{What the agents compute does not change.}{What the agents compute does not change.}}
The audit shows that the loss is not in computation. Within each of GPT-5.2 and GPT-5.1, the share of trajectories that compute the keyed value is similar under every delivery, while exposure and closure fall under TOC and Grep (Table~\ref{tab:master}, Build, Exposure, and Closure). Attributing the accuracy change to the stages gives the same picture, with closure the largest term for GPT-5.2 (Appendix~\ref{sec:appendix-contrasts}, Figure~\ref{fig:decomposition}).

\paragraph{The loss persists with the simulation held fixed.}
To separate delivery from computation, we take GPT-5.2's logged inline run and stop it at its first successful simulation, on \ReplayItems{} numeric items. From that point we continue the same history twice: once with the report inline, and once with the TOC (Appendix~\ref{sec:appendix-frozen-replay}). Both arms share the files, system prompt, reading tools, and remaining turns; they can read files and submit an answer but not rerun the simulator. Over three paired repeats, TOC changes accuracy by \ReplayDelta\,pp \ReplayCI{} (Table~\ref{tab:frozen-replay}). The loss is as large on the reports that fit inline and on the items whose frozen report contains the keyed value (Appendix~\ref{sec:appendix-frozen-replay}).

\begin{table}[t]
\centering
\footnotesize
\setlength{\tabcolsep}{4pt}
\caption{GPT-5.2 replay on \ReplayItems{} frozen checkpoints. Correct answers are shown over continuations per condition; $\Delta$ is TOC minus inline accuracy in pp. The mean uses three repeats of the same items.}
\label{tab:frozen-replay}
\begin{tabular}{lrrr}
\toprule
Repeat & Inline & TOC & $\Delta$ \\
\midrule
1 & \ReplayRunOneFullCorrect/\ReplayItems & \ReplayRunOneTOCCorrect/\ReplayItems & \ReplayRunOneDelta \\
2 & \ReplayRunTwoFullCorrect/\ReplayItems & \ReplayRunTwoTOCCorrect/\ReplayItems & \ReplayRunTwoDelta \\
3 & \ReplayRunThreeFullCorrect/\ReplayItems & \ReplayRunThreeTOCCorrect/\ReplayItems & \ReplayRunThreeDelta \\
\midrule
All & \ReplayFullCorrect/\ReplayPairs & \ReplayTOCCorrect/\ReplayPairs & \ReplayDelta \\
Mean (\%) & \ReplayFullAcc & \ReplayTOCAcc & \\
\bottomrule
\end{tabular}
\end{table}

\paragraph{\texorpdfstring{GPT-5.4 makes the same choice without answer options.}{GPT-5.4 makes the same choice without answer options.}}
Evidence substitution could be a reaction to options that differ only in the last digit. In a fill-in-the-blank version of the items, which asks for the numeric value itself, GPT-5.4 under TOC receives no part of the report on \fitbSkipN{} of \fitbVerN{} items and reads a file of its own in every such trajectory; \fitbMoreSkip{} of these answers carry more digits than the report prints (Appendix~\ref{sec:appendix-fitb}).

\paragraph{\texorpdfstring{GPT-5.4 shows the same behavior without a net loss.}{GPT-5.4 shows the same behavior without a net loss.}}
GPT-5.4 substitutes evidence as well: it skips the report in about four of ten TOC trajectories and chooses the truncated option in about half of the trap cases (Table~\ref{tab:master}, Trap). Its net change is nevertheless indistinguishable from zero, while its simulations compute the keyed value more often under TOC (Table~\ref{tab:master}, Build; Appendix~\ref{sec:appendix-contrasts}). A reading of this model's TOC errors by the benchmark's item author, made before our audit existed, names the same failure: in \expertPresentN{} of \expertErrN{} errors the right value was present in the agent's output and not used (Appendix~\ref{sec:appendix-expert}).

%% file: sections/table_master.tex
\begin{table}[t]
\centering
\footnotesize
\setlength{\tabcolsep}{5pt}
\caption{All agent conditions with trajectory logs, on the \VerN{} verified items (evidence use) and the \MechNumericN{} numeric items among them (trap and stages). \emph{Acc.}: accuracy, averaged over runs. \emph{Skip}: no part of the simulator's report reached the agent. \emph{Own}: the agent read one of its own selected-output files. \emph{Trap}: of trajectories that saw a value rounding to the key but truncating to a distractor, those that chose the distractor. \emph{Build}: the agent's simulation printed the keyed value. \emph{Exposure}: the value reached the agent, given that it was printed. \emph{Closure}: the answer was correct, given that the value was seen. Bold: the three values behind the main finding, in the TOC row of each GPT model (the report is skipped, the truncated option is chosen, and closure falls). TOC rows are the three logged reruns; \emph{TOC control} is the run made alongside the interventions. Claude Opus 4.6 has trajectory logs only for one rerun on the 33 long-output items and is not in this table; its accuracy is in Table~\ref{tab:main-results}.}
\label{tab:master}
\begin{tabular}{ll c c cc c ccc}
\toprule
& & & & \multicolumn{2}{c}{Evidence use (\%)} & & \multicolumn{3}{c}{Stage rates (\%)} \\
\cmidrule(lr){5-6} \cmidrule(lr){8-10}
Model & Condition & Runs & Acc. & Skip & Own & Trap & Build & Exposure & Closure \\
\midrule
\multirow{2}{*}{Sonnet 4.6} & TOC & 3 & 90.3\% & \nsSonnetTOC & \selSonnetTOC & \trapSonnetTOC & \pSonnetTOC & \seenSonnetTOC & \cSonnetTOC \\
 & Grep & 1 & \vSonnetGrep & \nsSonnetGrep & \selSonnetGrep & \trapSonnetGrep & \pSonnetGrep & \seenSonnetGrep & \cSonnetGrep \\
\midrule
\multirow{3}{*}{GPT-5.4} & Inline & 1 & \accGPTFiveFourRawLogged & \nsGPTFiveFourRaw & \selGPTFiveFourRaw & \trapGPTFiveFourRaw & \pGPTFiveFourRaw & \seenGPTFiveFourRaw & \cGPTFiveFourRaw \\
 & TOC & 3 & \mGPTFiveFourTOCAcc & \textbf{\nsGPTFiveFourTOC} & \selGPTFiveFourTOC & \textbf{\trapGPTFiveFourTOC} & \pGPTFiveFourTOC & \seenGPTFiveFourTOC & \textbf{\cGPTFiveFourTOC} \\
 & TOC+Report & 2 & \accGPTFiveFourRep & \nsGPTFiveFourRep & \selGPTFiveFourRep & \trapGPTFiveFourRep & \pGPTFiveFourRep & \seenGPTFiveFourRep & \cGPTFiveFourRep \\
\midrule
\multirow{7}{*}{GPT-5.2} & Inline & 1 & \mGPTFiveTwoRawAcc & \nsGPTFiveTwoRaw & \selGPTFiveTwoRaw & \trapGPTFiveTwoRaw & \pGPTFiveTwoRaw & \seenGPTFiveTwoRaw & \cGPTFiveTwoRaw \\
 & TOC & 3 & \mGPTFiveTwoTOCAcc & \textbf{\nsGPTFiveTwoTOC} & \selGPTFiveTwoTOC & \textbf{\trapGPTFiveTwoTOC} & \pGPTFiveTwoTOC & \seenGPTFiveTwoTOC & \textbf{\cGPTFiveTwoTOC} \\
 & Grep & 1 & \mGPTFiveTwoGrepAcc & \nsGPTFiveTwoGrep & \selGPTFiveTwoGrep & \trapGPTFiveTwoGrep & \pGPTFiveTwoGrep & \seenGPTFiveTwoGrep & \cGPTFiveTwoGrep \\
\addlinespace[2pt]
 & TOC control & 1 & \accGPTFiveTwoTOCOct & \nsGPTFiveTwoTOCOct & \selGPTFiveTwoTOCOct & \trapGPTFiveTwoTOCOct & \pGPTFiveTwoTOCOct & \seenGPTFiveTwoTOCOct & \cGPTFiveTwoTOCOct \\
 & TOC+Round & 2 & \accGPTFiveTwoRound & \nsGPTFiveTwoRound & \selGPTFiveTwoRound & \trapGPTFiveTwoRound & \pGPTFiveTwoRound & \seenGPTFiveTwoRound & \cGPTFiveTwoRound \\
 & TOC+Report & 2 & \accGPTFiveTwoRep & \nsGPTFiveTwoRep & \selGPTFiveTwoRep & \trapGPTFiveTwoRep & \pGPTFiveTwoRep & \seenGPTFiveTwoRep & \cGPTFiveTwoRep \\
 & Hybrid & 3 & \hGPTFiveTwoHybrid & \nsGPTFiveTwoHybrid & \selGPTFiveTwoHybrid & \trapGPTFiveTwoHybrid & \pGPTFiveTwoHybrid & \seenGPTFiveTwoHybrid & \cGPTFiveTwoHybrid \\
\midrule
\multirow{7}{*}{GPT-5.1} & Inline & 1 & \mGPTFiveOneRawAcc & \nsGPTFiveOneRaw & \selGPTFiveOneRaw & \trapGPTFiveOneRaw & \pGPTFiveOneRaw & \seenGPTFiveOneRaw & \cGPTFiveOneRaw \\
 & TOC & 3 & \mGPTFiveOneTOCAcc & \textbf{\nsGPTFiveOneTOC} & \selGPTFiveOneTOC & \textbf{\trapGPTFiveOneTOC} & \pGPTFiveOneTOC & \seenGPTFiveOneTOC & \textbf{\cGPTFiveOneTOC} \\
 & Grep & 1 & \mGPTFiveOneGrepAcc & \nsGPTFiveOneGrep & \selGPTFiveOneGrep & \trapGPTFiveOneGrep & \pGPTFiveOneGrep & \seenGPTFiveOneGrep & \cGPTFiveOneGrep \\
\addlinespace[2pt]
 & TOC control & 1 & \accGPTFiveOneTOCOct & \nsGPTFiveOneTOCOct & \selGPTFiveOneTOCOct & \trapGPTFiveOneTOCOct & \pGPTFiveOneTOCOct & \seenGPTFiveOneTOCOct & \cGPTFiveOneTOCOct \\
 & TOC+Round & 2 & \accGPTFiveOneRound & \nsGPTFiveOneRound & \selGPTFiveOneRound & \trapGPTFiveOneRound & \pGPTFiveOneRound & \seenGPTFiveOneRound & \cGPTFiveOneRound \\
 & TOC+Report & 2 & \accGPTFiveOneRep & \nsGPTFiveOneRep & \selGPTFiveOneRep & \trapGPTFiveOneRep & \pGPTFiveOneRep & \seenGPTFiveOneRep & \cGPTFiveOneRep \\
 & Hybrid & 3 & \hGPTFiveOneHybrid & \nsGPTFiveOneHybrid & \selGPTFiveOneHybrid & \trapGPTFiveOneHybrid & \pGPTFiveOneHybrid & \seenGPTFiveOneHybrid & \cGPTFiveOneHybrid \\
\bottomrule
\end{tabular}
\end{table}

%% file: sections/hybrid.tex
\section{Intervening on What Agents Read}
\label{sec:hybrid}
\label{sec:intervention}

\begin{figure}[t]
\centering
\includegraphics[width=0.55\linewidth]{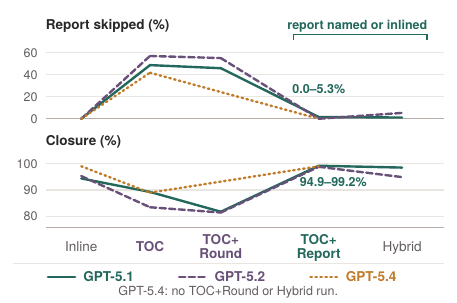}
\caption{Report skipping and closure under each condition. An instruction to round correctly (TOC+Round) does not reduce skipping and does not raise closure; naming the report (TOC+Report) or inlining reports that fit (Hybrid), the two bracketed conditions, removes the skipping and restores closure. TOC is the control run made alongside the interventions (GPT-5.4: its TOC reruns). TOC+Round and Hybrid were run for GPT-5.1 and GPT-5.2, so GPT-5.4's line joins TOC and TOC+Report directly.}
\label{fig:intervention}
\end{figure}

If the loss arises because agents answer from their own higher-precision files, changing which output they answer from should remove it without changing what they compute. We compare three variants with a fresh TOC run (TOC control): TOC+Report, Hybrid, and TOC+Round (Figure~\ref{fig:intervention} and Table~\ref{tab:master}). \textbf{TOC+Report} adds the following to the TOC tool description and to every tool result:
\begin{quote}\small
``IMPORTANT: The authoritative values are in result.out, printed at the precision the answer options use. Before answering, read the section of result.out that contains the quantity you need and take the value exactly as printed there. Do not answer from SELECTED\_OUTPUT or USER\_PUNCH files: they print unrounded values that can match a neighboring option.''
\end{quote}
\textbf{Hybrid} adds no instruction and inlines reports that fit within \RawCapChars{} characters.

\paragraph{The report instruction changes what agents read and removes the closure loss.}
With the report instruction, report skipping falls below 2\%, the agents rarely read their own output files, and no trajectory falls into the truncation trap (Figure~\ref{fig:intervention}, Table~\ref{tab:master}). Closure rises to about 99\%, while build stays within run-to-run variation. GPT-5.2 returns to its inline accuracy. GPT-5.1 gains about eight points over the TOC control but stays about five points below its inline accuracy. That gap is not in closure (Appendix~\ref{sec:appendix-contrasts}).

\paragraph{\texorpdfstring{Is the loss only a rounding error?}{Is the loss only a rounding error?}}
The example in Figure~\ref{fig:method} invites a simpler explanation: the agents may just be bad at rounding. The two explanations predict different things. If the problem is rounding, an instruction to round correctly should help. If the problem is which file the agent reads, naming the report should help, and a rounding instruction should not. To test this, \textbf{TOC+Round} replaces the instruction with one that gives the rounding rule and names no file:
\begin{quote}\small
``IMPORTANT: The answer options give each value at a fixed precision. If a value you read has more digits than the options, round it to the options' precision to the nearest value; do not truncate digits. Choose the option that equals the correctly rounded value.''
\end{quote}
The rounding instruction does not help. Agents still skip the report about as often as under TOC, closure is about 82\%, no higher than under the TOC control, and they still choose the truncated option in about seven of ten trap cases, although told not to truncate (Table~\ref{tab:rounding-test}). Accuracy changes by \dRoundOctGPTFiveTwoP{} (GPT-5.2) and \dRoundOctGPTFiveOneP\,pp (GPT-5.1) against the TOC control, with no consistent gain. The same agents answer correctly when the report is in front of them and they answer from it: with inline output, closure is 94--99\% (Table~\ref{tab:master}). Truncation is also not the whole loss: with every truncated answer counted as correct, GPT-5.1 still changes by \truncForgivenOneP\,pp (Table~\ref{tab:truncation}). The loss is therefore not a rounding error on its own. The agents truncate because they answer from a file that prints more digits, and they stop when they are directed to the value that is already printed at the right precision.

\begin{table}[!ht]
\centering
\small
\setlength{\tabcolsep}{5pt}
\caption{Is the loss only a rounding error? The three TOC conditions differ only in the instruction. \emph{Skip}: share of trajectories in which no part of the report reached the agent (\%). \emph{Trap}: of the trajectories that saw a value rounding to the keyed value but truncating to another option, those that chose the truncated option. \emph{Closure}: share of trajectories that answered correctly, among those that saw the keyed value (\%). The rounding instruction leaves all three about where they were, or worse; naming the report changes all three (bold; 0/0: no trajectory saw such a value, because the agents read the report). Values are taken from Table~\ref{tab:master}.}
\label{tab:rounding-test}
\begin{tabular}{llccccccc}
\toprule
 & & \multicolumn{3}{c}{GPT-5.2} & & \multicolumn{3}{c}{GPT-5.1} \\
\cmidrule(lr){3-5}\cmidrule(lr){7-9}
Condition & The agent is told & Skip & Trap & Closure & & Skip & Trap & Closure \\
\midrule
TOC control & nothing & \nsGPTFiveTwoTOCOct & \trapGPTFiveTwoTOCOct & \cGPTFiveTwoTOCOct & & \nsGPTFiveOneTOCOct & \trapGPTFiveOneTOCOct & \cGPTFiveOneTOCOct \\
TOC+Round & how to round & \nsGPTFiveTwoRound & \trapGPTFiveTwoRound & \cGPTFiveTwoRound & & \nsGPTFiveOneRound & \trapGPTFiveOneRound & \cGPTFiveOneRound \\
TOC+Report & which file to read & \textbf{\nsGPTFiveTwoRep} & \textbf{\trapGPTFiveTwoRep} & \textbf{\cGPTFiveTwoRep} & & \textbf{\nsGPTFiveOneRep} & \textbf{\trapGPTFiveOneRep} & \textbf{\cGPTFiveOneRep} \\
\bottomrule
\end{tabular}
\end{table}

\paragraph{The report instruction also lifts GPT-5.4.}
With the report instruction, GPT-5.4 reaches \accGPTFiveFourRep{} in both runs, about eight points above both its TOC runs and its inline runs, with intervals that exclude zero. Report skipping falls to zero and closure returns to 99\%, while the build gain of retrieval remains, so the instruction removes the loss from evidence substitution while keeping what retrieval adds.

\paragraph{Inlining what fits has the same effect.}
Over three runs of Hybrid, report skipping nearly disappears and closure returns to about its inline level (Table~\ref{tab:master}). Accuracy rises over the TOC control by \dHybOctGPTFiveTwoP{} (GPT-5.2) and \dHybOctGPTFiveOneP\,pp (GPT-5.1). Relative to the inline runs, GPT-5.2 is within about 1\,pp and GPT-5.1 is again about five points lower, with the residual outside closure.

\paragraph{\texorpdfstring{Summary.}{Summary.}}
Both interventions restore closure without changing build, and for GPT-5.2 the paired replay shows the loss with computation fixed. So the closure loss comes from which file the agent answers from. Hybrid needs no knowledge of the tool; the instruction keeps retrieval but must name the canonical output.

%% file: sections/discussion.tex
\section{Discussion}
\label{sec:discussion}

\paragraph{Retrieval adds a choice of evidence.}
The two deliveries make the same report obtainable: every number in the inline report is one read away, and for GPT-5.2, with the simulation held fixed, the loss remains on reports that fit inline. What differs is which output the agent answers from. In our setting, retrieval turns reading the report into a choice of evidence. Inline, one source is in view; under retrieval, the report competes with the files the agent wrote itself, and models that choose badly lose accuracy. The instruction works because it makes that choice for them.

\paragraph{\texorpdfstring{When the substitution costs accuracy.}{When the substitution costs accuracy.}}
The substitution costs accuracy here because the two outputs print the same value differently and the items require the report's precision. This is the condition under which the mechanism applies: a tool with a canonical output, and an agent that can define an output of its own whose form need not match the form in which answers are checked.

\paragraph{Implications for harness design.}
In our setting, final accuracy does not show where the loss arises, and for GPT-5.4 it does not show the substitution at all. Stating which output is authoritative removed the substitution; inlining outputs that fit had the same effect and needs no knowledge of the tool.

%% file: sections/related_work.tex
\section{Related Work}
\label{sec:related}

\paragraph{Delivering tool output to agents.}
The agent--computer interface, including how files are viewed and searched, changes what software-engineering agents achieve \citep{yang2024sweagent}. Long tool responses in context degrade function calling \citep{kate2025longfunceval}, and harnesses respond by truncating, masking, or compressing observations \citep{lindenbauer2025complexity,kang2025acon} or by writing them to files that the agent opens or searches \citep{abramovich2025enigma,anthropic2025context,katz2026dynamic,curme2026deepagents}. File-based delivery helps when the material is far larger than the context window \citep{cao2026codingagents}. For material that fits, controlled comparisons report effects of both signs: \citet{sen2026grep} find that delivering search results as files lowers accuracy in nine of ten harness--model pairs, and state that they lack trace-level attribution; \citet{mcmillan2026structured} finds that delivering a database schema as files slightly helps frontier models and hurts open-weight models. On the items we use, \citet{zhang2026phreeqcmcq} report that a table-of-contents interface lowers the accuracy of GPT-5.1 and GPT-5.2 and attribute this to navigation ability. We reproduce that net effect on verified items and revise the explanation: we locate the loss within individual trajectories and test the location by intervention.

\paragraph{\texorpdfstring{Whether agents use the evidence available to them.}{Whether agents use the evidence available to them.}}
Models do not always use context that contains what they need \citep{joren2025sufficient}, and agents often see relevant content without acting on it \citep{englander2026agents}. Closest to the behavior we report, \citet{roh2026before} show from tool-call logs that agentic-RAG controllers often answer without reading the passages they retrieved, and that an environment rule that rejects such answers raises accuracy. Agents also rarely request the rest of an over-long tool response \citep{petrova2026paginate} or recall observations that were moved to external storage \citep{fan2026harness}. In these studies an agent that skips the evidence has no other source of the answer. In ours it has one: a second output of the same tool call, which holds the correct value in another representation. Studies of source preference find that models favor generated or self-authored text over retrieved text when the two disagree \citep{tan2024blinded,druck2026rag}; our two outputs agree in value, and the cost arises when the value is transcribed. Rounding to significant digits is unreliable in isolation \citep{yang2025numbercookbook}; we observe truncation in place of rounding when a value is copied from a source with more digits.

\paragraph{Locating agent failures.}
Agent failures are usually attributed to a step or an agent: by human or model judges \citep{zhang2025whoandwhen}, by an attributor trained on counterfactual replays \citep{zhang2026agentracer}, or by replaying the agent with a proposed repair \citep{lin2026reflect}. Other diagnostics separate evidence that was never obtained from evidence that was obtained and not used: RAGChecker with model-based claim checking \citep{ru2024ragchecker}, \citet{liu2026diagnosing} with human relevance labels for search agents, and \citet{xie2026bayes} with rule-scored hits for messages between agents. \citet{xie2026bayes} decompose communication loss into externalization, absorption, and action closure; our exposure stage is closest to their externalization, and our closure stage merges their absorption and action closure, which we do not separate. Deterministic scoring of recorded tool traces \citep{flynt2026groundeval} and replay from a frozen state \citep{pakhomov2026attributing} have also been proposed. A deterministic simulator adds a label for the stage that precedes these: re-executing the inputs the agent itself wrote shows whether it computed the target value at all. Agents for scientific software are usually studied on one package and graded on outcomes: OpenFOAM cases that run \citep{yue2026foamagent,chen2024metaopenfoam}, COMSOL targets \citep{mudur2025feabench}, or molecular-dynamics subtasks \citep{campbell2025mdcrow}; SUPER checks whether expected outputs appear in executed cells \citep{bogin2024super}.

%% file: sections/limitations.tex
\section{Limitations}
\label{sec:limitations}

\paragraph{Models.}
All models are closed and accessed through their APIs. Evidence substitution appears in all three GPT models and rarely in Claude Sonnet 4.6, the Claude model with full trajectory logs; we do not determine which model property drives this difference.

\paragraph{Decoding and prompts.}
The Claude agents ran at temperature 0 without extended thinking, and the GPT agents at API defaults with no reasoning-effort setting, so we do not know whether the effect persists with explicit reasoning. We test one wording of each instruction.

%% file: sections/conclusion.tex
\section{Conclusion}

On \VerN{} PHREEQC items, three GPT models that receive the simulator's report through retrieval often leave it unopened, answer from an output file of their own design, and truncate its value. Re-executing every logged trajectory shows that they compute the keyed value about as often as before; for GPT-5.2, holding the simulation fixed confirms that the loss comes from the delivery. Naming the authoritative output, or inlining reports that fit, removes the substitution; asking for correct rounding does not. Re-executing the agents' own inputs in a deterministic tool showed where the answers were lost, which final accuracy did not.

%% file: appendix/item_verification.tex
\section{Item Verification}
\label{sec:appendix-verification}

\paragraph{Procedure.}
Every item in the \PoolN-item pool was re-executed in PHREEQC 3.8.6 with the frozen \texttt{phreeqc.dat}. For the 9 setups where the item author's original PHREEQC input was available we ran it unchanged, adding only output statements; for the remaining setups we encoded the item text literally with PHREEQC defaults (mmol/kgw, 25\,$^\circ$C, pe 4 unless stated). Inputs were never adjusted toward the key; where nothing matched, a small number of documented alternative readings (step numbering, units, charge balance) were tried and recorded. For redox-sensitive quantities in systems without a redox couple, we re-ran with a tighter convergence tolerance and alternative step-size and kinetic-integrator settings, and marked an item unstable if the printed value changed at the digit that separates the options. An item is retained only if its printed value matches the keyed option exactly and uniquely and is stable. Model results played no part in this process. Table~\ref{tab:exclusions} lists the excluded items; the item author reviewed all of them and agreed that each is defective as written.

\begin{table}[h]
\centering
\footnotesize
\setlength{\tabcolsep}{3pt}
\caption{Excluded items by reason (dataset item IDs).}
\label{tab:exclusions}
\begin{tabular}{p{0.38\linewidth}cp{0.42\linewidth}}
\toprule
Reason & $n$ & Items \\
\midrule
Incorrect key & 1 & 22 \\
Unstable redox-derived value & 17 & 1, 8, 48, 108, 115, 118, 119, 125, 128, 130, 133, 136, 138, 140, 145, 166, 192 \\
Missing question sentence & 3 & 4, 37, 78 \\
Malformed option or species name & 2 & 3, 186 \\
Asks for moles, key is molality & 3 & 54, 64, 66 \\
Wrong process or step in stem & 3 & 56 (precipitated, not dissolved); 5, 6 (key matches step 48) \\
Ambiguous location (exchanger vs.\ cell) & 4 & 151, 153, 159, 160 \\
Missing units in setup & 9 & 163--165, 167--172 \\
\midrule
Total & 42 & \\
\bottomrule
\end{tabular}
\end{table}

\paragraph{Results by item set.}
Table~\ref{tab:item-set-comparison} compares the full pool, the \VerN{} verified items used in the main text, and a stricter subset of \StrictN{} items that also drops the 24 retained items whose setup text under-specifies the reference input. Accuracies rise when defective items are removed, because models were often marked wrong on items whose key or wording was defective, but the ordering of models and the pattern of TOC versus inline (no detectable difference for the Claude models and GPT-5.4, a loss for GPT-5.1 and GPT-5.2) are the same in all three.

\begin{table}[t]
\centering
\footnotesize
\setlength{\tabcolsep}{3pt}
\caption{Headline-run results on the full pool (\PoolN), the verified items (\VerN), and the strict subset (\StrictN).}
\label{tab:item-set-comparison}
\begin{tabular}{l ccc ccc ccc}
\toprule
& \multicolumn{3}{c}{Agent-TOC acc.} & \multicolumn{3}{c}{Agent-Inline acc.} & \multicolumn{3}{c}{TOC $-$ Inline} \\
\cmidrule(lr){2-4} \cmidrule(lr){5-7} \cmidrule(lr){8-10}
Model & \PoolN & \VerN & \StrictN & \PoolN & \VerN & \StrictN & \PoolN & \VerN & \StrictN \\
\midrule
Claude Opus 4.6   & \fOpusTOC & \vOpusTOC & \sOpusTOC & \fOpusRaw & \vOpusRaw & \sOpusRaw & \fOpusTOCMinusRaw & \vOpusTOCMinusRaw & \sOpusTOCMinusRaw \\
Claude Sonnet 4.6 & \fSonnetTOC & \vSonnetTOC & \sSonnetTOC & \fSonnetRaw & \vSonnetRaw & \sSonnetRaw & \fSonnetTOCMinusRaw & \vSonnetTOCMinusRaw & \sSonnetTOCMinusRaw \\
GPT-5.4 & \fGPTFiveFourTOC & \vGPTFiveFourTOC & \sGPTFiveFourTOC & \fGPTFiveFourRaw & \vGPTFiveFourRaw & \sGPTFiveFourRaw & \fGPTFiveFourTOCMinusRaw & \vGPTFiveFourTOCMinusRaw & \sGPTFiveFourTOCMinusRaw \\
GPT-5.2 & \fGPTFiveTwoTOC & \vGPTFiveTwoTOC & \sGPTFiveTwoTOC & \fGPTFiveTwoRaw & \vGPTFiveTwoRaw & \sGPTFiveTwoRaw & \fGPTFiveTwoTOCMinusRaw & \vGPTFiveTwoTOCMinusRaw & \sGPTFiveTwoTOCMinusRaw \\
GPT-5.1 & \fGPTFiveOneTOC & \vGPTFiveOneTOC & \sGPTFiveOneTOC & \fGPTFiveOneRaw & \vGPTFiveOneRaw & \sGPTFiveOneRaw & \fGPTFiveOneTOCMinusRaw & \vGPTFiveOneTOCMinusRaw & \sGPTFiveOneTOCMinusRaw \\
Gemini 3 Flash & \fGeminiThreeFlashTOC & \vGeminiThreeFlashTOC & \sGeminiThreeFlashTOC & \fGeminiThreeFlashRaw & \vGeminiThreeFlashRaw & \sGeminiThreeFlashRaw & -- & \vGeminiThreeFlashTOCMinusRaw & -- \\
\bottomrule
\end{tabular}
\end{table}

%% file: appendix/conditions.tex
\section{Prompts, Tool Results, and Conditions}
\label{sec:appendix-intervention}

\paragraph{Prompts and tool definitions.}
Each full agent receives a system prompt, the item as its user prompt, and four tool definitions through the API (Figure~\ref{fig:prompts}); Grep adds a fifth, and the definitions are otherwise the same in every condition. Across conditions, the system prompt changes only in its \texttt{execute\_phreeqc} line (Figure~\ref{fig:execute-lines}) and in the mode named in its last line. The no-tool baselines send the item text alone with one of two system prompts (Figure~\ref{fig:prompts}).

\begin{figure}[t]
\begin{minipage}[t]{0.49\textwidth}
\vspace{0pt}
\noindent\fbox{\begin{minipage}{\dimexpr\linewidth-2\fboxsep-2\fboxrule\relax}
{\small\rmfamily\bfseries System prompt (full agents, TOC condition)}\\[3pt]
\scriptsize\ttfamily\raggedright
\#~Role\\
You are an expert geochemist and PHREEQC simulation specialist.\\
Your goal is to answer multiple-choice geochemistry questions (A/B/C/D) by writing and executing PHREEQC simulations.\\[3pt]
\#~Tools\\
- list\_file(path): List files/directories in the workspace.\\
- read\_file(path, start\_line?, end\_line?): Read a text file. Large files are auto-truncated; use start\_line/end\_line for specific sections.\\
- write\_file(path, new\_content): Create or overwrite a text file.\\
- execute\_phreeqc(input\_path): Run PHREEQC on an input file. Returns metadata and a TOC for result.out. File contents are NOT included \textemdash{} use read\_file to inspect results.\\[3pt]
\#~Rules\\
- All file operations must use relative paths within the workspace.\\
- Do NOT include a DATABASE line in PHREEQC input files (the database is provided externally).\\
- A database reference (database\_reference.txt) is available in your workspace listing valid species, minerals, gases, and built-in RATES blocks. Consult it before writing input files.\\
- Base your final answer on PHREEQC output, not memory. Do not guess.\\
- If your computed value does not match any option, your input is likely wrong \textemdash{} fix and re-run rather than picking the closest option.\\
- Call only one tool per turn.\\
- Write your final answer to final\_answer.txt with exact content: \textless{}\textless{}\textless{} X \textgreater{}\textgreater{}\textgreater{} followed by a newline.\\[3pt]
\#~Context mode for this run\\
The execute\_phreeqc tool is configured in \textasciigrave{}toc\textasciigrave{} mode. Use the payload it returns according to the tool contract above.
\end{minipage}}
\end{minipage}\hfill
\begin{minipage}[t]{0.49\textwidth}
\vspace{0pt}
\noindent\fbox{\begin{minipage}{\dimexpr\linewidth-2\fboxsep-2\fboxrule\relax}
{\small\rmfamily\bfseries User prompt (full agents)}\\[3pt]
\scriptsize\ttfamily\raggedright
Answer the multiple-choice question below.\\[3pt]
Question:\\
\textrm{\textit{[item text with options A--D]}}\\[3pt]
Use tools as needed for PHREEQC / quantitative work.
\end{minipage}}
\par\vspace{4pt}
\noindent\fbox{\begin{minipage}{\dimexpr\linewidth-2\fboxsep-2\fboxrule\relax}
{\small\rmfamily\bfseries Tool definitions (API function schemas)}\\[3pt]
\scriptsize\ttfamily\raggedright
write\_file(path, new\_content): Create or overwrite a text file in the workspace.\\[2pt]
read\_file(path, start\_line?, end\_line?): Read a text file. Optionally specify start\_line/end\_line (1-based) for a range.\\[2pt]
list\_file(path?): List files and directories in the workspace.\\[2pt]
execute\_phreeqc(input\_path): Run PHREEQC on an input file. Returns execution status, stdout/stderr, output file paths, and the result payload configured for this run's context mode.\\[2pt]
grep\_file(path, pattern, max\_matches?, context\_lines?, case\_sensitive?) \textrm{\textit{[Grep only]}}: Search a text file line-by-line with a regular expression (case-insensitive by default). Returns matching line numbers with surrounding context lines. Use read\_file with start\_line/end\_line to inspect a region around a match.
\end{minipage}}
\par\vspace{4pt}
\noindent\fbox{\begin{minipage}{\dimexpr\linewidth-2\fboxsep-2\fboxrule\relax}
{\small\rmfamily\bfseries System prompt (Direct baseline)}\\[3pt]
\scriptsize\ttfamily\raggedright
You are a geochemistry expert. Answer the multiple-choice question. Your entire response must be EXACTLY \textless{}\textless{}\textless{} X \textgreater{}\textgreater{}\textgreater{} where X is A, B, C, or D. No explanation, no reasoning, no other text. ONLY \textless{}\textless{}\textless{} X \textgreater{}\textgreater{}\textgreater{}.
\end{minipage}}
\par\vspace{4pt}
\noindent\fbox{\begin{minipage}{\dimexpr\linewidth-2\fboxsep-2\fboxrule\relax}
{\small\rmfamily\bfseries System prompt (CoT baseline)}\\[3pt]
\scriptsize\ttfamily\raggedright
You are a geochemistry expert. Solve the multiple-choice question by reasoning step by step, but do not use tools. End your response with the final answer in the exact form \textless{}\textless{}\textless{} X \textgreater{}\textgreater{}\textgreater{}, where X is A, B, C, or D.
\end{minipage}}
\end{minipage}
\caption{Prompts and tool definitions, verbatim. Left: the system prompt of the full agents, shown for TOC; other conditions replace its \texttt{execute\_phreeqc} line (Figure~\ref{fig:execute-lines}) and the mode name. Right: their user prompt; the tool definitions passed through the API, with parameters abridged (``?'': optional), the same in every condition except that \texttt{grep\_file} is offered only in Grep; and the system prompts of the no-tool baselines, whose user prompt is the item text alone.}
\label{fig:prompts}
\end{figure}

\begin{figure}[t]
\noindent\fbox{\begin{minipage}{\dimexpr\linewidth-2\fboxsep-2\fboxrule\relax}
{\small\rmfamily\bfseries The \texttt{execute\_phreeqc} line of the system prompt, by condition}\\[3pt]
\footnotesize\raggedright
\textbf{Inline:} ``Run PHREEQC on an input file. Returns metadata and raw result.out content inline, truncated if very large. Use read\_file only if you need to inspect an additional output file or re-read a specific portion.''\\[3pt]
\textbf{TOC:} ``Run PHREEQC on an input file. Returns metadata and a TOC for result.out. File contents are NOT included --- use read\_file to inspect results.''\\[3pt]
\textbf{Grep:} ``Run PHREEQC on an input file. Returns metadata only (status, output file paths and sizes, result.out line count). File contents are NOT included --- locate evidence with grep\_file, then read specific line ranges with read\_file.'' followed by ``grep\_file(path, pattern, max\_matches?, context\_lines?, case\_sensitive?): Search a text file line-by-line with a regular expression (case-insensitive by default). Returns matching line numbers with a few lines of surrounding context. Use it to find sections, species, or numeric fields in result.out, then read\_file with start\_line/end\_line to inspect the full region.'' and ``NOTE: In this mode, read\_file on a large file WITHOUT start\_line/end\_line returns no content. Always grep\_file first to find line numbers, then read the range you need.''\\[3pt]
\textbf{Hybrid:} ``Run PHREEQC on an input file. Returns metadata plus the raw result.out content inline when it is small enough. For a large result.out it returns a TOC instead (section names with starting line numbers) and no content; in that case use read\_file with start\_line/end\_line to read the sections you need.''\\[3pt]
\textbf{TOC+Round:} the TOC line followed by ``IMPORTANT: The answer options give each value at a fixed precision. If a value you read has more digits than the options, round it to the options' precision to the nearest value; do not truncate digits. Choose the option that equals the correctly rounded value.''\\[3pt]
\textbf{TOC+Report:} the TOC line followed by ``IMPORTANT: The authoritative values are in result.out, printed at the precision the answer options use. Before answering, read the section of result.out that contains the quantity you need and take the value exactly as printed there. Do not answer from SELECTED\_OUTPUT or USER\_PUNCH files: they print unrounded values that can match a neighboring option.''
\end{minipage}}
\caption{The condition-specific \texttt{execute\_phreeqc} line of the system prompt, verbatim. TOC+Report and TOC+Round also repeat their instruction in every tool result.}
\label{fig:execute-lines}
\end{figure}

The inline description discourages additional reads, and the retrieval descriptions require them; part of the difference in reading behavior between inline and retrieval conditions may therefore be induced by the descriptions rather than by the return value alone; the paired replay, whose arms share one system prompt, still shows the loss (Appendix~\ref{sec:appendix-frozen-replay}). The TOC+Report instruction both names the authoritative file and gives a reason to prefer it; TOC+Round gives only the rounding rule and names no file.

\paragraph{Tool results.}
Every \texttt{execute\_phreeqc} call returns the exit status, the last 4{,}000 characters of PHREEQC's standard output and error, and the names and sizes of the files the run created. Inline adds the report, cut to its first and last 50{,}000 characters with a truncation marker when it is longer than 100{,}000. TOC adds an index of the report's dash-delimited section headers with their starting lines, at most 50 entries (the first and last 25 when a report has more), and the hint ``Use read\_file with start\_line/end\_line to read specific sections.'' In the headline TOC runs and the TOC reruns the index was empty for reports longer than 100{,}000 characters, so on those items all four TOC runs use one separate run with the index built; all other runs build it for every report. Hybrid adds the report when it fits and otherwise the index and the report length; Grep adds only the report's line count. TOC+Report and TOC+Round return the TOC payload with their instruction appended to the hint, so the agent sees the instruction in the system prompt and again in every \texttt{execute\_phreeqc} result. With a line range, \texttt{read\_file} returns the requested lines; without one it returns a file of up to 20{,}000 characters whole and otherwise its first and last 80 lines (no content in Grep).

\begin{table}[t]
\centering
\small
\begin{tabular}{lcccccc}
\toprule
Model & Inline & TOC & Grep & Hybrid & TOC+Report & TOC+Round \\
\midrule
Opus 4.6 & 1 & 1 & -- & -- & -- & -- \\
Sonnet 4.6 & 1 & 4 & 1 & -- & -- & -- \\
GPT-5.4 & 2 & 4 & -- & -- & 2 & -- \\
GPT-5.2 & 2 & 4+1 & 1 & 3 & 2 & 2 \\
GPT-5.1 & 2 & 4+1 & 1 & 3 & 2 & 2 \\
Gemini 3 Flash & 1 & 1 & -- & -- & -- & -- \\
\bottomrule
\end{tabular}
\caption{Full-agent runs per model and condition on the \VerN{} items. Each model has one headline inline run and one headline TOC run; the other TOC runs are reruns, and on the 33 items whose report exceeds 100k characters the four TOC runs share one run with a fixed index. +1: the TOC control run of the interventions, with the index built for every report size.}
\label{tab:runs}
\end{table}

\paragraph{\texorpdfstring{Accuracy per model and no-tool baselines.}{Accuracy per model and no-tool baselines.}}
Table~\ref{tab:main-results} gives the headline accuracy of every model with and without the simulator. With Direct prompting, accuracy is \vGPTFiveFourOneShot--\vOpusOneShot, at most about 10\,pp above the \VerMajority{} majority-label rate. Simulator access adds \vToolGainMin--\vToolGainMax\,pp over the better no-tool baseline for the five Claude and GPT models.

\begin{table}[t]
\centering
\footnotesize
\setlength{\tabcolsep}{4pt}
\caption{Accuracy on the \VerN{} verified items. Direct, CoT, Agent-TOC, and Agent-Inline are the headline runs; unparseable or missing answers count as incorrect. The last two columns report the primary contrast: per-item accuracy under TOC minus inline, with every run of each condition pooled per item (Table~\ref{tab:runs}; the pool includes the TOC control run of Section~\ref{sec:hybrid}), a 95\% item-bootstrap interval, and a Holm-corrected sign-flip permutation $p$ over the five models.}
\label{tab:main-results}
\begin{tabular}{lcccccc}
\toprule
Model & Direct & CoT & Agent-TOC & Agent-Inline & TOC $-$ inline, pooled (pp) & $p_{\text{Holm}}$ \\
\midrule
Claude Opus 4.6 & \vOpusOneShot & \vOpusCoT & \vOpusTOC & \vOpusRaw & \dTOCRawOpus & \pHolmTOCRawOpus \\
Claude Sonnet 4.6 & \vSonnetOneShot & \vSonnetCoT & \vSonnetTOC & \vSonnetRaw & \dTOCRawSonnet & \pHolmTOCRawSonnet \\
GPT-5.4 & \vGPTFiveFourOneShot & \vGPTFiveFourCoT & \vGPTFiveFourTOC & \vGPTFiveFourRaw & \dTOCRawGPTFiveFour & \pHolmTOCRawGPTFiveFour \\
GPT-5.2 & \vGPTFiveTwoOneShot & \vGPTFiveTwoCoT & \vGPTFiveTwoTOC & \vGPTFiveTwoRaw & \dTOCRawGPTFiveTwo & \pHolmTOCRawGPTFiveTwo \\
GPT-5.1 & \vGPTFiveOneOneShot & \vGPTFiveOneCoT & \vGPTFiveOneTOC & \vGPTFiveOneRaw & \dTOCRawGPTFiveOne & \pHolmTOCRawGPTFiveOne \\
\bottomrule
\end{tabular}
\end{table}

\paragraph{\texorpdfstring{Run details.}{Run details.}}
The TOC reruns cover Sonnet and the three GPT models, Grep covers Sonnet, GPT-5.2, and GPT-5.1, the interventions cover the GPT models, and the fill-in-the-blank analysis covers one TOC run of GPT-5.4; the other combinations were not run or not logged, and their table cells are empty. The full-agent conditions were run at different times. The four TOC runs share one run on the 33 long-output items, and the decomposition uses one logged inline run per model. The paired replay covers GPT-5.2 and starts from checkpoints of its inline run. The full-agent conditions differ in the system-prompt line that describes the tool's return value, and the two instruction conditions also repeat their instruction in every tool result; the paired replay uses the same prompt in both arms.

\paragraph{The intervention runs.}
The intervention and TOC control runs used the same harness, item file, and step budget, with the index built for every report size. TOC+Round reached \accGPTFiveTwoRoundRuns{} for GPT-5.2 and \accGPTFiveOneRoundRuns{} for GPT-5.1. TOC+Report reached \accGPTFiveTwoRepRuns{} for GPT-5.2, \accGPTFiveOneRepRuns{} for GPT-5.1, and \accGPTFiveFourRepRuns{} for GPT-5.4 in its two runs; GPT-5.4's logged inline run reached \accGPTFiveFourRawLogged. With the instruction, GPT-5.4 defined a selected-output file in only \defGPTFiveFourRep\% of trajectories (\defGPTFiveFourTOC\% under TOC), and \chanRepGPTFiveFourRep\% of its correct numeric answers that saw the keyed value took it from the report (\chanRepGPTFiveFourTOC\% under TOC). Of the correct numeric answers that saw the keyed value, the share that saw it in the report rather than only in the agent's own selected-output files was \chanRepGPTFiveTwoTOCOct\% (GPT-5.2) and \chanRepGPTFiveOneTOCOct\% (GPT-5.1) in the TOC control, \chanRepGPTFiveTwoRep\% and \chanRepGPTFiveOneRep\% with the instruction, \chanRepGPTFiveTwoHybrid\% and \chanRepGPTFiveOneHybrid\% under Hybrid, and \chanRepGPTFiveTwoRaw\% and \chanRepGPTFiveOneRaw\% in the inline runs. Accuracy gains under the instruction therefore come from answers taken from the report, not from correctly rounded selected-output values. \tokHybridText{} Mean input tokens per item under the instruction were \tokGPTFiveTwoTOCOct{} (control) and \tokGPTFiveTwoRep{} (instruction) for GPT-5.2, and \tokGPTFiveOneTOCOct{} and \tokGPTFiveOneRep{} for GPT-5.1.

\begin{table}[t]
\centering
\footnotesize
\setlength{\tabcolsep}{3pt}
\caption{Accuracy over runs on the \VerN{} verified items. TOC ranges cover the headline run and three reruns, which share one run on the 33 long-output items; Inline lists the headline and the logged run (GPT-5.1, GPT-5.2, GPT-5.4); Hybrid ranges cover three runs. Tokens are mean input tokens per item from the runs with full logs; full-set logs are unavailable for Opus and for Sonnet inline. $^\dagger$Mostly a change in behavior unrelated to the condition (Appendix~\ref{sec:appendix-intervention}).}
\label{tab:toc-raw}
\begin{tabular}{l cc c c c c ccc}
\toprule
& \multicolumn{2}{c}{Retrieval} & & Inline & & & \multicolumn{3}{c}{Input tokens per item} \\
\cmidrule(lr){2-3} \cmidrule(lr){5-5} \cmidrule(lr){8-10}
Model & TOC & Grep & & Inline & & Hybrid & TOC & Inline & Hybrid \\
\midrule
Opus 4.6   & \vOpusTOC & -- & & \vOpusRaw & & -- & \tokOpusTOC & -- & -- \\
Sonnet 4.6 & \vSonnetTOCRunMin--\vSonnetTOCRunMax & \vSonnetGrep & & \vSonnetRaw & & -- & \tokSonnetTOC & -- & -- \\
GPT-5.4    & \vGPTFiveFourTOCRunMin--\vGPTFiveFourTOCRunMax & -- & & \vGPTFiveFourRaw, \accGPTFiveFourRawLogged & & -- & \tokGPTFiveFourTOC & \tokGPTFiveFourRaw & -- \\
GPT-5.2    & \vGPTFiveTwoTOCRunMin--\vGPTFiveTwoTOCRunMax & \vGPTFiveTwoGrep & & \vGPTFiveTwoRaw, \vGPTFiveTwoRawRepeat & & \hGPTFiveTwoHybridMin--\hGPTFiveTwoHybridMax & \tokGPTFiveTwoTOC & \tokGPTFiveTwoRaw & \tokGPTFiveTwoHybrid \\
GPT-5.1    & \vGPTFiveOneTOCRunMin--\vGPTFiveOneTOCRunMax & \vGPTFiveOneGrep & & \vGPTFiveOneRaw, \vGPTFiveOneRawRepeat & & \hGPTFiveOneHybridMin--\hGPTFiveOneHybridMax & \tokGPTFiveOneTOC & \tokGPTFiveOneRaw & \tokGPTFiveOneHybrid \\
\bottomrule
\end{tabular}
\end{table}

%% file: appendix/failure_localization.tex
\section{Failure Localization}
\label{sec:appendix-localization}

\paragraph{Trajectories.}
We analyze every full-agent run with complete logs on the \VerN{} verified items: the three TOC reruns per model for Sonnet, GPT-5.4, GPT-5.2, and GPT-5.1 (items whose report exceeds 100k characters come from the run with the fixed index, as in the headline numbers), the logged inline and Grep runs, the Hybrid, TOC+Report, TOC+Round, and TOC control runs, the GPT-5.4 TOC+Report and logged inline runs, the Opus run with the fixed index (33 items), and the Gemini 3 Flash runs. This gives \MechTrajN{} distinct trajectories. Each workspace was matched to its item through the question text in the log.

\paragraph{Produced and seen.}
For every \texttt{execute\_phreeqc} call we rebuilt the input from the agent's preceding \texttt{write\_file} events and re-ran it with the same PHREEQC 3.8.6 binary and database (\MechCallsN{} calls, \MechInputsN{} distinct inputs). The \emph{produced} corpus is every output of these executions; the \emph{seen} corpus is every tool result the agent received that carries simulator output (inline payloads, \texttt{read\_file} results on the report or selected-output files, and search matches). A keyed value counts as present if a printed number equals the option at its precision, or rounds to it when printed with more digits; numbers on a rounding midpoint match both neighbors; for quantities described as dissolved, precipitated, or changed, the sign is ignored; specific-conductance options match only on conductance lines; the echo of the input in the report is excluded. A trajectory is labeled \emph{build} if the keyed value was never produced, \emph{exposure} if it was produced but never seen, and \emph{closure} if it was seen but the answer was wrong; trajectories without a gradable answer and wrong answers on non-numeric items are counted separately.

\begin{table}[t]
\centering
\footnotesize
\setlength{\tabcolsep}{2.4pt}
\caption{Where accuracy is lost, on numeric items. Left: stage rates ($p$ = the agent's own simulation printed the keyed value; $s$ = the value reached the agent, given produced; $c$ = correct, given seen). Right: Shapley attribution of each condition's accuracy change relative to the same model's logged inline run, in pp (95\% item-bootstrap intervals in Table~\ref{tab:shapley-ci}); \emph{unsupported} covers answers given without the keyed value in view. TOC averages the three logged reruns. The intervention conditions are in Table~\ref{tab:intervention}.}
\label{tab:mechanism}
\begin{tabular}{ll ccc c ccccc}
\toprule
& & \multicolumn{3}{c}{Stage rates (\%)} & & \multicolumn{5}{c}{Change vs.\ inline (pp)} \\
\cmidrule(lr){3-5} \cmidrule(lr){7-11}
Model & Condition & $p$ & $s$ & $c$ & & Total & Build & Exposure & Closure & Unsupp. \\
\midrule
Sonnet 4.6 & TOC & \pSonnetTOC & \seenSonnetTOC & \cSonnetTOC & & -- & -- & -- & -- & -- \\
\midrule
\multirow{2}{*}{GPT-5.4} & Inline & \pGPTFiveFourRaw & \seenGPTFiveFourRaw & \cGPTFiveFourRaw & & \multicolumn{5}{c}{(reference)} \\
 & TOC    & \pGPTFiveFourTOC & \seenGPTFiveFourTOC & \cGPTFiveFourTOC & & \shTotFourTOCP & \textbf{\shBldFourTOCP} & \shAccFourTOCP & \shCloFourTOCP & \shUnsFourTOCP \\
\midrule
\multirow{3}{*}{GPT-5.2} & Inline & \pGPTFiveTwoRaw & \seenGPTFiveTwoRaw & \cGPTFiveTwoRaw & & \multicolumn{5}{c}{(reference)} \\
 & TOC    & \pGPTFiveTwoTOC & \seenGPTFiveTwoTOC & \cGPTFiveTwoTOC & & \shTotTwoTOCP & \shBldTwoTOCP & \shAccTwoTOCP & \textbf{\shCloTwoTOCP} & \shUnsTwoTOCP \\
 & Grep   & \pGPTFiveTwoGrep & \seenGPTFiveTwoGrep & \cGPTFiveTwoGrep & & \shTotTwoGrepP & \shBldTwoGrepP & \shAccTwoGrepP & \shCloTwoGrepP & \shUnsTwoGrepP \\
\midrule
\multirow{3}{*}{GPT-5.1} & Inline & \pGPTFiveOneRaw & \seenGPTFiveOneRaw & \cGPTFiveOneRaw & & \multicolumn{5}{c}{(reference)} \\
 & TOC    & \pGPTFiveOneTOC & \seenGPTFiveOneTOC & \cGPTFiveOneTOC & & \shTotOneTOCP & \shBldOneTOCP & \textbf{\shAccOneTOCP} & \textbf{\shCloOneTOCP} & \shUnsOneTOCP \\
 & Grep   & \pGPTFiveOneGrep & \seenGPTFiveOneGrep & \cGPTFiveOneGrep & & \shTotOneGrepP & \shBldOneGrepP & \shAccOneGrepP & \shCloOneGrepP & \shUnsOneGrepP \\
\bottomrule
\end{tabular}
\end{table}

\begin{table}[t]
\centering
\scriptsize
\setlength{\tabcolsep}{3pt}
\caption{Shapley attribution of each condition's accuracy change relative to the same model's logged inline run (pp, numeric items), with 95\% item-bootstrap intervals (1{,}000 resamples, items resampled jointly for both conditions). TOC: the three reruns; TOC (control), TOC+Report, and Hybrid: the intervention runs.}
\label{tab:shapley-ci}
\begin{tabular}{llccccc}
\toprule
Model & Condition & Total & Build & Exposure & Closure & Unsupported \\
\midrule
GPT-5.2 & TOC          & \shTotTwoTOC & \shBldTwoTOC & \shAccTwoTOC & \shCloTwoTOC & \shUnsTwoTOC \\
        & Grep         & \shTotTwoGrep & \shBldTwoGrep & \shAccTwoGrep & \shCloTwoGrep & \shUnsTwoGrep \\
        & TOC (control) & \shTotTwoTOCOct & \shBldTwoTOCOct & \shAccTwoTOCOct & \shCloTwoTOCOct & \shUnsTwoTOCOct \\
        & TOC+Report   & \shTotTwoRep & \shBldTwoRep & \shAccTwoRep & \shCloTwoRep & \shUnsTwoRep \\
        & Hybrid       & \shTotTwoHyb & \shBldTwoHyb & \shAccTwoHyb & \shCloTwoHyb & \shUnsTwoHyb \\
\midrule
GPT-5.4 & TOC          & \shTotFourTOC & \shBldFourTOC & \shAccFourTOC & \shCloFourTOC & \shUnsFourTOC \\
        & TOC+Report   & \shTotFourRep & \shBldFourRep & \shAccFourRep & \shCloFourRep & \shUnsFourRep \\
\midrule
GPT-5.1 & TOC          & \shTotOneTOC & \shBldOneTOC & \shAccOneTOC & \shCloOneTOC & \shUnsOneTOC \\
        & Grep         & \shTotOneGrep & \shBldOneGrep & \shAccOneGrep & \shCloOneGrep & \shUnsOneGrep \\
        & TOC (control) & \shTotOneTOCOct & \shBldOneTOCOct & \shAccOneTOCOct & \shCloOneTOCOct & \shUnsOneTOCOct \\
        & TOC+Report   & \shTotOneRep & \shBldOneRep & \shAccOneRep & \shCloOneRep & \shUnsOneRep \\
        & Hybrid       & \shTotOneHyb & \shBldOneHyb & \shAccOneHyb & \shCloOneHyb & \shUnsOneHyb \\
\bottomrule
\end{tabular}
\end{table}

\paragraph{Reliability.}
For numeric trajectories with a last execution available for comparison, re-running that execution reproduces its stored \texttt{result.out} in \MechValidOK{} of \MechValidN{} cases; the rest differ only in line order. The keyed value appears in our reference output for all \MechNumericN{} numeric items; for \MechUnambigN{} of them no distractor value appears anywhere in that output, and the main results hold on this subset (GPT-5.2 TOC: closure \shCloTwoTOCUnamb\,pp, exposure \shAccTwoTOCUnamb\,pp; GPT-5.1 TOC: closure \shCloOneTOCUnamb\,pp, exposure \shAccOneTOCUnamb\,pp). Because matching is numeric, a printed number could also match the keyed value on a line that reports a different quantity. On the \onceN{} numeric items whose keyed value occurs only once in the reference report, where this is least likely, the TOC terms are closure \onceCloTwo{} and exposure \onceAccTwo{} for GPT-5.2, closure \onceCloOne{} and exposure \onceAccOne{} for GPT-5.1, and closure \onceCloFour{} and exposure \onceAccFour{} for GPT-5.4 (pp, 95\% item-bootstrap intervals). Among correct numeric answers of the Claude and GPT agents, the keyed value was seen in 69--95\% of trajectories; the remainder are mostly nearest-option picks when no option value was produced, which we report as unsupported answers. Re-execution moved only \MechReexecMovedN{} trajectories from build to exposure. Of the \MechBuildFailuresN{} build failures of the Claude and GPT agents, \MechBuildNoOptionN{} produced none of the four option values, and re-running the agents' inputs with PHREEQC's full default printing recovered the keyed value in \MechBuildDefaultRecoveredN.

\paragraph{Manual audit.}
Independently of the automatic labels, we read 68 trajectories: 12 TOC losses each for GPT-5.1 and GPT-5.2 and 6 Grep losses for GPT-5.2 (items wrong under retrieval but correct inline, each read together with its inline trajectory, 60 in all), and 8 Sonnet TOC failures. Seven of the 24 TOC losses were caused by the retrieval channel (the inline run had the value in context, while the TOC run never saw it or misread its own selected output), two were mistakes with the value in hand, and fifteen were build failures, seven of which the inline run also had and answered correctly by choosing the nearest option. Losses selected on the outcome are therefore dominated by build failures that churn between runs and net out, which is why we rely on the decomposition rather than on item-level losses. On the 58 numeric trajectories in this sample, the automatic stage (build, exposure, closure, or correct with the keyed value in view) agrees with the manual one in 52 cases. Five of the six disagreements count a matching number printed for another reaction step or solution as the keyed value, which moves a build failure into exposure or closure (or, once, a lucky answer into a supported one); the sixth misses a value the agent saw only as a logarithm. The sample was selected on outcomes, so this agreement rate is indicative rather than a population estimate. Navigating to the wrong report section occurred once. All eight Sonnet failures were build failures. Representative excerpts: ``gives pH = 2.75065, which rounds to 2.750'' (key 2.751); ``Talc: 23.333 mol \dots\ The mineral that dissolved the most is Talc'' (final amount read as amount reacted; the inline run with the same input saw the change column).

\paragraph{Answers without the value in view.}
GPT-5.1 declines (``none of the options'') in \declGPTFiveOneTOC\% of TOC trajectories and \declGPTFiveOneRaw\% inline; in 19 of its 22 distinct numeric TOC non-commits the keyed value was never produced. Inline, the same model tends to pick the nearest option instead, which is correct about a third of the time. About half of GPT-5.2's distinct TOC non-commits (8 of 15) are step-budget exhaustion.

\begin{table}[t]
\centering
\footnotesize
\setlength{\tabcolsep}{3.2pt}
\caption{Intervention runs on the \VerN{} verified items. $\Delta$: per-item accuracy difference from the TOC control run (GPT-5.4: from all four TOC runs; pp). Skip, Own, and Digits as in Table~\ref{tab:evidence}; \emph{Def.}: the agent defined a selected-output file. Stage rates and the closure term (Shapley, pp relative to the logged inline run) as in Table~\ref{tab:mechanism}. The inline rows are the logged inline runs, shown for reference; GPT-5.4's TOC row averages its three TOC reruns.}
\label{tab:intervention}
\begin{tabular}{ll c cc cccc ccc c}
\toprule
& & & \multicolumn{2}{c}{Accuracy} & \multicolumn{4}{c}{Evidence use (\%)} & \multicolumn{3}{c}{Stage rates (\%)} & Closure \\
\cmidrule(lr){4-5} \cmidrule(lr){6-9} \cmidrule(lr){10-12}
Model & Condition & Runs & \% & $\Delta$ & Skip & Def. & Own & Digits & $p$ & $s$ & $c$ & term \\
\midrule
\multirow{5}{*}{GPT-5.2} & Inline & 1 & \mGPTFiveTwoRawAcc & -- & \nsGPTFiveTwoRaw & \defGPTFiveTwoRaw & \selGPTFiveTwoRaw & \unrGPTFiveTwoRaw & \pGPTFiveTwoRaw & \seenGPTFiveTwoRaw & \cGPTFiveTwoRaw & (ref.) \\
 & TOC control & 1 & \accGPTFiveTwoTOCOct & -- & \nsGPTFiveTwoTOCOct & \defGPTFiveTwoTOCOct & \selGPTFiveTwoTOCOct & \unrGPTFiveTwoTOCOct & \pGPTFiveTwoTOCOct & \seenGPTFiveTwoTOCOct & \cGPTFiveTwoTOCOct & \shCloTwoTOCOctP \\
 & TOC+Report & 2 & \accGPTFiveTwoRep & \dRepOctGPTFiveTwoP & \textbf{\nsGPTFiveTwoRep} & \defGPTFiveTwoRep & \selGPTFiveTwoRep & \unrGPTFiveTwoRep & \pGPTFiveTwoRep & \seenGPTFiveTwoRep & \textbf{\cGPTFiveTwoRep} & \shCloTwoRepP \\
 & TOC+Round & 2 & \accGPTFiveTwoRound & \dRoundOctGPTFiveTwoP & \nsGPTFiveTwoRound & \defGPTFiveTwoRound & \selGPTFiveTwoRound & \unrGPTFiveTwoRound & \pGPTFiveTwoRound & \seenGPTFiveTwoRound & \cGPTFiveTwoRound & \shCloTwoRoundP \\
 & Hybrid & 3 & \hGPTFiveTwoHybrid & \dHybOctGPTFiveTwoP & \nsGPTFiveTwoHybrid & \defGPTFiveTwoHybrid & \selGPTFiveTwoHybrid & \unrGPTFiveTwoHybrid & \pGPTFiveTwoHybrid & \seenGPTFiveTwoHybrid & \cGPTFiveTwoHybrid & \shCloTwoHybP \\
\midrule
\multirow{3}{*}{GPT-5.4} & Inline & 1 & \accGPTFiveFourRawLogged & -- & \nsGPTFiveFourRaw & \defGPTFiveFourRaw & \selGPTFiveFourRaw & \unrGPTFiveFourRaw & \pGPTFiveFourRaw & \seenGPTFiveFourRaw & \cGPTFiveFourRaw & (ref.) \\
 & TOC & 3 & \mGPTFiveFourTOCAcc & -- & \nsGPTFiveFourTOC & \defGPTFiveFourTOC & \selGPTFiveFourTOC & \unrGPTFiveFourTOC & \pGPTFiveFourTOC & \seenGPTFiveFourTOC & \cGPTFiveFourTOC & \shCloFourTOCP \\
 & TOC+Report & 2 & \accGPTFiveFourRep & \dRepTOCGPTFiveFourP & \textbf{\nsGPTFiveFourRep} & \defGPTFiveFourRep & \selGPTFiveFourRep & \unrGPTFiveFourRep & \pGPTFiveFourRep & \seenGPTFiveFourRep & \textbf{\cGPTFiveFourRep} & \shCloFourRepP \\
\midrule
\multirow{5}{*}{GPT-5.1} & Inline & 1 & \mGPTFiveOneRawAcc & -- & \nsGPTFiveOneRaw & \defGPTFiveOneRaw & \selGPTFiveOneRaw & \unrGPTFiveOneRaw & \pGPTFiveOneRaw & \seenGPTFiveOneRaw & \cGPTFiveOneRaw & (ref.) \\
 & TOC control & 1 & \accGPTFiveOneTOCOct & -- & \nsGPTFiveOneTOCOct & \defGPTFiveOneTOCOct & \selGPTFiveOneTOCOct & \unrGPTFiveOneTOCOct & \pGPTFiveOneTOCOct & \seenGPTFiveOneTOCOct & \cGPTFiveOneTOCOct & \shCloOneTOCOctP \\
 & TOC+Report & 2 & \accGPTFiveOneRep & \dRepOctGPTFiveOneP & \textbf{\nsGPTFiveOneRep} & \defGPTFiveOneRep & \selGPTFiveOneRep & \unrGPTFiveOneRep & \pGPTFiveOneRep & \seenGPTFiveOneRep & \textbf{\cGPTFiveOneRep} & \shCloOneRepP \\
 & TOC+Round & 2 & \accGPTFiveOneRound & \dRoundOctGPTFiveOneP & \nsGPTFiveOneRound & \defGPTFiveOneRound & \selGPTFiveOneRound & \unrGPTFiveOneRound & \pGPTFiveOneRound & \seenGPTFiveOneRound & \cGPTFiveOneRound & \shCloOneRoundP \\
 & Hybrid & 3 & \hGPTFiveOneHybrid & \dHybOctGPTFiveOneP & \nsGPTFiveOneHybrid & \defGPTFiveOneHybrid & \selGPTFiveOneHybrid & \unrGPTFiveOneHybrid & \pGPTFiveOneHybrid & \seenGPTFiveOneHybrid & \cGPTFiveOneHybrid & \shCloOneHybP \\
\bottomrule
\end{tabular}
\end{table}

\begin{table}[t]
\centering
\footnotesize
\setlength{\tabcolsep}{1.7pt}
\caption{Which evidence the agents use (\%, except Trap counts). \emph{Skip}: no part of \texttt{result.out} reached the agent. \emph{Own}: the agent read one of its own selected-output files. \emph{Digits}: of trajectories that saw the keyed value, those that saw it only with more digits than the report. \emph{Trap}: of trajectories that saw a value rounding to the key but truncating to a distractor, those that chose the distractor (each trajectory once). \emph{Corr.\,P / Corr.\,D}: accuracy when the keyed value was seen as the report prints it / only with more digits.}
\label{tab:evidence}
\begin{tabular}{llcccccc}
\toprule
Model & Cond. & Skip & Own & Digits & Trap & Corr.\,P & Corr.\,D \\
\midrule
Sonnet 4.6 & TOC & \nsSonnetTOC & \selSonnetTOC & \unrSonnetTOC & \trapSonnetTOC & \medExSonnet & \medUnSonnet \\
GPT-5.4 & TOC & \textbf{\nsGPTFiveFourTOC} & \selGPTFiveFourTOC & \unrGPTFiveFourTOC & \trapGPTFiveFourTOC & \medExGPTFiveFour & \medUnGPTFiveFour \\
GPT-5.2 & TOC & \textbf{\nsGPTFiveTwoTOC} & \selGPTFiveTwoTOC & \unrGPTFiveTwoTOC & \textbf{\trapGPTFiveTwoTOC} & \medExGPTFiveTwo & \medUnGPTFiveTwo \\
GPT-5.1 & TOC & \textbf{\nsGPTFiveOneTOC} & \selGPTFiveOneTOC & \unrGPTFiveOneTOC & \textbf{\trapGPTFiveOneTOC} & \medExGPTFiveOne & \medUnGPTFiveOne \\
\midrule
GPT-5.4 & Inline & \nsGPTFiveFourRaw & \selGPTFiveFourRaw & \unrGPTFiveFourRaw & \trapGPTFiveFourRaw & \medExGPTFiveFourRaw & \medUnGPTFiveFourRaw \\
GPT-5.2 & Inline & \nsGPTFiveTwoRaw & \selGPTFiveTwoRaw & \unrGPTFiveTwoRaw & \trapGPTFiveTwoRaw & \cGPTFiveTwoRaw & -- \\
GPT-5.1 & Inline & \nsGPTFiveOneRaw & \selGPTFiveOneRaw & \unrGPTFiveOneRaw & \trapGPTFiveOneRaw & \cGPTFiveOneRaw & -- \\
\bottomrule
\end{tabular}
\end{table}

%% file: appendix/contrasts.tex
\section{\texorpdfstring{Detailed Contrasts}{Detailed Contrasts}}
\label{sec:appendix-contrasts}

This appendix gives the detailed contrasts, intervals, and counts behind Sections~\ref{sec:main-results}--\ref{sec:hybrid}. Table~\ref{tab:alternatives} lists the comparisons and what each isolates.

\begin{table}[t]
\centering
\footnotesize
\setlength{\tabcolsep}{3pt}
\begin{tabular}{p{0.40\linewidth}p{0.55\linewidth}}
\toprule
Comparison & What it isolates \\
\midrule
Re-execution of all \MechCallsN{} simulator calls & What the agent computed: the build term of the accuracy change stays within 2\,pp for GPT-5.1 and GPT-5.2 (\S\ref{sec:mechanism}) \\
Paired replay from the same executed simulation & Delivery alone: \ReplayDelta\,pp for GPT-5.2 (\S\ref{sec:mechanism}) \\
Replay on the \ReplayFitN{} reports that fit inline & Delivery without report length: same loss (\S\ref{sec:mechanism}) \\
Keyword search instead of the section index & Retrieval without simulator-specific structure: same cost (\S\ref{sec:main-results}) \\
Fill-in-the-blank items & The choice of evidence without answer options (\S\ref{sec:mechanism}) \\
TOC+Round against TOC+Report & How to round against which output to read: only the latter helps (\S\ref{sec:hybrid}) \\
Reruns pooled per item; setups resampled and left out one at a time & Run-to-run and setup-to-setup variation (this appendix) \\
Manual labels and expert review & The automatic stage labels: agreement on 52 of 58 trajectories (App.~\ref{sec:appendix-localization},~\ref{sec:appendix-expert}) \\
Re-verification of answer keys & The items: \VerN{} of 200 retained (\S\ref{sec:items}) \\
\bottomrule
\end{tabular}
\caption{The comparisons in the paper and what each one isolates.}
\label{tab:alternatives}
\end{table}

\paragraph{Pooled TOC-versus-inline contrast (Section~\ref{sec:main-results}).}
Pooling every run per item, TOC changes accuracy relative to inline output by \dTOCRawGPTFiveOne{} for GPT-5.1 and \dTOCRawGPTFiveTwo{} for GPT-5.2 (Figure~\ref{fig:net-cost} and Table~\ref{tab:main-results}; the pool includes the TOC control run of Section~\ref{sec:hybrid}). After Holm correction over the five models, the GPT-5.1 difference is significant ($p=\pHolmTOCRawGPTFiveOne$) and the GPT-5.2 difference is not ($p=\pHolmTOCRawGPTFiveTwo$; unadjusted $\pTOCRawGPTFiveTwo$); with the simulation held fixed, GPT-5.2's TOC cost is \ReplayDelta\,pp \ReplayCI{} (Section~\ref{sec:mechanism}). Resampling the 19 setups instead of items gives intervals of \setupCIOne{} (GPT-5.1) and \setupCITwo{} (GPT-5.2), and leaving out any one setup moves the estimates only within \looOne{} and \looTwo\,pp. For both models, the two inline runs score higher than every TOC run and the Grep run (Table~\ref{tab:toc-raw}). The Grep run, which uses no PHREEQC section structure, shows a similar cost.

The pooled differences of Claude Opus 4.6, Claude Sonnet 4.6, and GPT-5.4 are \dTOCRawOpusP, \dTOCRawSonnetP, and \dTOCRawGPTFiveFourP\,pp, with intervals that include zero; for GPT-5.4 the stage terms below have opposite signs.

\begin{figure}[t]
\centering
\includegraphics[width=0.55\linewidth]{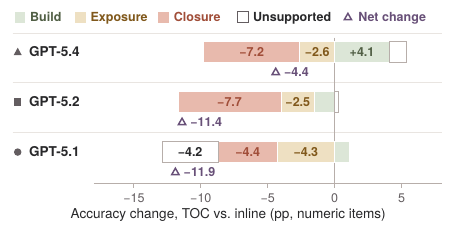}
\caption{Where the accuracy change under TOC comes from, relative to each model's logged inline run on the \MechNumericN{} numeric items (Shapley attribution, pp; stage rates and 95\% intervals in Tables~\ref{tab:mechanism} and~\ref{tab:shapley-ci}). Exposure and closure are negative in all three models; in GPT-5.4 a gain in build offsets part of them.}
\label{fig:decomposition}
\end{figure}

\paragraph{Exposure and closure (Section~\ref{sec:mechanism}).}
For GPT-5.2 under TOC, closure accounts for \shCloTwoTOCP{} of the \shTotTwoTOCP\,pp change on numeric items, with an interval that excludes zero, and exposure for \shAccTwoTOCP\,pp. For GPT-5.1 the loss splits about evenly between exposure and closure, each with an interval that excludes zero (Tables~\ref{tab:mechanism} and~\ref{tab:shapley-ci}). Grep shows the same pattern. These totals are relative to the single logged inline run and cover numeric items only, so they differ from the pooled contrast in Table~\ref{tab:main-results}. Figure~\ref{fig:decomposition} shows the attribution for the three GPT models.

Under TOC, GPT-5.2 and GPT-5.1 skip the report in \nsGPTFiveTwoTOC\% and \nsGPTFiveOneTOC\% of trajectories, and Sonnet in \nsSonnetTOC\% (Table~\ref{tab:evidence}). Trap cases make up about two thirds of the TOC trajectories that saw the keyed value and still answered wrongly. Answers were nearly always correct when the keyed value was seen as the report prints it, and correct in \medUnGPTFiveTwo\% (GPT-5.2) and \medUnGPTFiveOne\% (GPT-5.1) of the trajectories that saw it only with more digits. When the agent received both the report's form of the keyed value and a value that truncates to another option, which happens mostly when the report is inline, GPT-5.1 and GPT-5.2 answered correctly in \bothTrapGPT{} trajectories (inline and Hybrid runs) and Sonnet in \bothTrapSonnet{} (TOC runs).

When the keyed value was computed but never seen, the agent had skipped the report in \expoSkipGPTFiveOne{} cases for GPT-5.1 and \expoSkipGPTFiveTwo{} for GPT-5.2. Under TOC, GPT-5.2 and GPT-5.1 take more steps than inline (\effStepsTOC{} versus \effStepsRaw{} steps), and one of the 30 retrieval losses that we read by hand involved a wrong report section.

GPT-5.4's two inline runs scored \vGPTFiveFourRaw{} and \accGPTFiveFourRawLogged; the second has trajectory logs and is the baseline of the decomposition. GPT-5.4 skips the report in \nsGPTFiveFourTOC\% of TOC trajectories (Table~\ref{tab:evidence}). Relative to its logged inline run, TOC costs it about ten points in exposure and closure (\shAccFourTOCP{} and \shCloFourTOCP\,pp), about as much as GPT-5.2, and it gains \shBldFourTOC\,pp in build (Table~\ref{tab:mechanism}). On this basis (numeric items, three TOC reruns against the logged inline run) the net change is \shTotFourTOC\,pp, and pooled over all runs and items it is \dTOCRawGPTFiveFourP\,pp. Neither is distinguishable from zero, while its exposure and closure terms are both negative with intervals that exclude zero.

\paragraph{Interventions (Section~\ref{sec:hybrid}).}
With the report instruction, report skipping falls to \nsGPTFiveTwoRep\% (GPT-5.2) and \nsGPTFiveOneRep\% (GPT-5.1), and 99\% of the correct numeric answers that saw the keyed value took it from the report. Closure rises to \cGPTFiveTwoRep\% and \cGPTFiveOneRep\%, and build is unchanged within run-to-run variation (Table~\ref{tab:intervention}). Accuracy rises over all five TOC runs by \dRepTOCGPTFiveTwoP{} and \dRepTOCGPTFiveOneP\,pp (\dRepTOCBoth{} averaged over the two models) and over the TOC control by \dRepOctGPTFiveTwoP{} and \dRepOctGPTFiveOneP\,pp, a comparison limited by one control run per model. GPT-5.1 remains below its inline accuracy (\dRepRawGPTFiveOne), with the residual in the build, exposure, and unsupported terms.

Under the rounding instruction, the agents choose the truncated option in \trapGPTFiveTwoRound{} (GPT-5.2) and \trapGPTFiveOneRound{} (GPT-5.1) trap cases. Accuracy changes by \dRoundTOCGPTFiveTwoP{} and \dRoundTOCGPTFiveOneP\,pp relative to all TOC runs and by \dRoundOctGPTFiveTwoP{} and \dRoundOctGPTFiveOneP\,pp relative to the TOC control. TOC+Report exceeds TOC+Round by \dRepRoundGPTFiveTwo{} and \dRepRoundGPTFiveOne.

With the report instruction, GPT-5.4 reaches \accGPTFiveFourRep{} in both runs, \dRepTOCGPTFiveFour{} above its TOC runs and \dRepRawGPTFiveFour{} above its inline runs. Report skipping falls to \nsGPTFiveFourRep\% and closure rises to \cGPTFiveFourRep\%, and the build gain under TOC remains (\shBldFourRepP\,pp relative to inline).

Over three runs, Hybrid raises accuracy over the TOC control by \dHybOctGPTFiveTwoP{} (GPT-5.2) and \dHybOctGPTFiveOneP\,pp (GPT-5.1). Relative to the inline runs, GPT-5.2 changes by \dHybRawGPTFiveTwoP\,pp and GPT-5.1 by \dHybRawGPTFiveOneP\,pp, with GPT-5.1's residual outside closure.

\paragraph{\texorpdfstring{How much of the loss is truncation.}{How much of the loss is truncation.}}
In all \truncFellN{} TOC trajectories in which a GPT model chose the option to which a received value truncates, that value came from the agent's own file and never from the report. The models chose the truncated option in \truncLitPct\% of the cases in which that option is a literal prefix of the printed value (\truncLit) and in \truncConvPct\% of those in which the two are printed in different notations (\truncConv); the two groups cover different items. Table~\ref{tab:truncation} recomputes the TOC-minus-inline difference on the runs with trajectory logs after counting as correct every answer that chose the option to which a received value truncates. The comparison uses the three TOC reruns and the TOC control against the logged inline run of each model, so its first column differs slightly from the pooled contrast in Table~\ref{tab:main-results}. For GPT-5.2 and GPT-5.4 the remaining difference is within about 3\,pp of zero; for GPT-5.1 it is \truncForgivenOneP\,pp with an interval that excludes zero, and lies mostly outside closure (Figure~\ref{fig:decomposition}).

\begin{table}[t]
\centering
\footnotesize
\setlength{\tabcolsep}{4pt}
\begin{tabular}{lcc}
\toprule
Model & As graded & Truncation forgiven \\
\midrule
GPT-5.4 & \truncGradedFour & \truncForgivenFour \\
GPT-5.2 & \truncGradedTwo & \truncForgivenTwo \\
GPT-5.1 & \truncGradedOne & \truncForgivenOne \\
\bottomrule
\end{tabular}
\caption{Accuracy under TOC minus inline on the logged runs (pp, 95\% item-bootstrap intervals), as graded and with truncated answers counted as correct (\emph{truncation forgiven}).}
\label{tab:truncation}
\end{table}

%% file: appendix/frozen_replay.tex
\section{Paired Replay with Computation Held Fixed}
\label{sec:appendix-frozen-replay}

\paragraph{Checkpoints.}
We take the first execution returning exit code zero in GPT-5.2's logged inline run, before its result reaches the model. Of the \MechNumericN{} numeric items, \ReplayItems{} have such a checkpoint; the remaining six have no successful execution. Selection uses the execution status, independently of the original final answer or whether the output contains the keyed value. We reconstruct the visible history and workspace by replaying the preceding file writes and PHREEQC calls in order, including earlier failed executions. For the \ReplayFitN{} untruncated reports, re-execution matches the recorded report after normalizing path and timing lines and separator rules; the other \ReplayLongN{} original returns contain only the capped head and tail. Both delivery arms use the same reconstructed files, verified by hashes.

\paragraph{Paired continuations.}
Only the pending execution result differs: the inline arm returns the report with the usual \RawCapChars{}-character cap, whereas TOC returns its section index, built at every report size. The arms share a common system prompt, the visible history, file-reading tools, and the remaining budget of $24-s$ assistant turns, where $s$ is the execution's turn. They can list or read the frozen files and submit \texttt{final\_answer.txt}; simulation files remain fixed. We run \ReplayRuns{} continuations per item and condition with GPT-5.2's API defaults and a 4{,}096-completion-token cap per call. Item order and the order of the two arms within each pair are shuffled with a fixed seed (20261011), and the paired arms run consecutively. The first repeat was inspected before the two additional repeats were run. These \ReplayBranches{} continuations are counted separately from the full-agent trajectories in Appendix~\ref{sec:appendix-localization}.

\paragraph{Aggregation.}
Table~\ref{tab:frozen-replay} in the main text gives the counts per repeat. For each item we average the three paired accuracy differences, then average across the \ReplayItems{} items. A missing submitted answer counts as incorrect; one TOC continuation in repeat 2 exhausts its remaining turns. We compute a percentile 95\% interval with 50{,}000 paired-item bootstrap resamples (seed 20261011), retaining all three repeats of each sampled item. The mean difference is \ReplayDelta\,pp, with interval \ReplayCI. As a sensitivity check for shared setups, we resample the \ReplaySetups{} setups with all their items and repeats, retaining the item-weighted mean; the interval is \ReplayClusterCI.

\paragraph{Report size and available values.}
The \ReplayFitN{} frozen reports that fit inline yield \ReplayFitFullAcc\% inline and \ReplayFitTOCAcc\% under TOC (\ReplayFitDelta\,pp). For the \ReplayLongN{} longer reports the corresponding accuracies are \ReplayLongFullAcc\% and \ReplayLongTOCAcc\%. Using the matching rule in Appendix~\ref{sec:appendix-localization}, the keyed value is present in the frozen report for \ReplayKeyN{} items; accuracy on these items is \ReplayKeyFullAcc\% inline and \ReplayKeyTOCAcc\% under TOC. The fixed outputs therefore retain the same value while the delivery choice changes how often the agent answers correctly.

%% file: appendix/fitb_evidence.tex
\section{\texorpdfstring{Evidence Use without Answer Options}{Evidence Use without Answer Options}}
\label{sec:appendix-fitb}

The fill-in-the-blank version of the items keeps each question stem, removes the four options, and asks for the numeric value, which is graded against the key with a per-item absolute tolerance (for most items just under half a unit of the key's last printed digit). It covers the 182 items of the pool whose answer is a number without a unit; \fitbVerN{} of them are verified items. The agent, its tools, and the TOC interface are those of Section~\ref{sec:harness}; the system prompt differs in the lines that state the task and the answer format. We analyze the one logged run, GPT-5.4 under TOC, which answers \fitbTOC{} of the \fitbVerN{} items within tolerance, against \fitbDirect{} and \fitbCoT{} for Direct and CoT prompting.

Table~\ref{tab:fitb-evidence} splits the trajectories by whether any part of \texttt{result.out} reached the agent. The agent skips the report on \fitbSkipN{} items (\fitbSkipPct\%), and each time it reads a file that its own simulation wrote. \fitbMoreSkip{} of these answers carry more significant digits than the key, which has the precision of the report, so the printed answer itself shows that it was not taken from the report. Because the grader accepts a value within tolerance, answering from the more precise file is not penalized here as it is with options one digit apart, and the two groups differ little in accuracy. The run shows that the choice of evidence does not depend on the options; it does not measure a cost of retrieval, since no inline fill-in-the-blank run with logs is available for comparison.

\begin{table}[t]
\centering
\footnotesize
\setlength{\tabcolsep}{4pt}
\begin{tabular}{lcc}
\toprule
 & Skipped & Read \\
\midrule
Trajectories & \fitbSkipN & \fitbReadN \\
Read its own output file & \fitbOwnSkip & \fitbOwnRead \\
More digits than the key & \fitbMoreSkip & \fitbMoreRead \\
Within tolerance & \fitbCorrSkip & \fitbCorrRead \\
\bottomrule
\end{tabular}
\caption{Evidence use of GPT-5.4 under TOC on the \fitbVerN{} verified fill-in-the-blank items (one logged run). \emph{Skipped}: no part of \texttt{result.out} reached the agent; \emph{Read}: some part did.}
\label{tab:fitb-evidence}
\end{table}

%% file: appendix/gemini_step_budget.tex
\section{Gemini 3 Flash Outcomes under TOC and Inline Output}
\label{sec:appendix-gemini-step-budget}

On the \VerN{} items, Gemini 3 Flash Preview reaches \vGeminiThreeFlashTOC{} under TOC and \vGeminiThreeFlashRaw{} inline, against \vGeminiThreeFlashOneShot{} and \vGeminiThreeFlashCoT{} with Direct and CoT prompting. Most of its failures are missing answers rather than wrong ones (Table~\ref{tab:appendix-gemini-step-budget}). Under TOC, 80 trajectories use up the 24-step budget without committing an answer, against 33 inline; navigating the index takes a median of 8 \texttt{read\_file} calls per item, against 2 inline. In another 30 (TOC) and 40 (inline) trajectories the agent stops after giving its answer in a chat message instead of writing the answer file. When it does commit a parseable answer, it is correct about as often under both conditions (44 of 46 and 79 of 85), so the gain from inline delivery comes from committing more often, not from answering better. Counts are from the logs of the headline run for each condition.

\begin{table}[t]
\centering
\small
\begin{tabular}{lcc}
\toprule
Outcome (\VerN{} items) & TOC & Inline \\
\midrule
Correct & 44 & 79 \\
Wrong & 2 & 6 \\
Answer only in a chat message & 30 & 40 \\
Answer file unparseable & 1 & 0 \\
Step budget exhausted, no answer & 80 & 33 \\
Log incomplete & 1 & 0 \\
\midrule
Accuracy when committed & 95.7\% & 92.9\% \\
Median \texttt{read\_file} calls & 8 & 2 \\
Median steps & 24 & 17 \\
\bottomrule
\end{tabular}
\caption{Gemini 3 Flash outcomes on the \VerN{} items in the headline TOC and inline runs. \emph{Accuracy when committed}: correct among correct and wrong answers.}
\label{tab:appendix-gemini-step-budget}
\end{table}

%% file: appendix/expert_review.tex
\section{\texorpdfstring{Human Expert Baseline and Expert Review of Failures}{Human Expert Baseline and Expert Review of Failures}}
\label{sec:appendix-expert}

\paragraph{\texorpdfstring{Human expert worksheet.}{Human expert worksheet.}}
A PHREEQC user with eight years of experience who did not write the items and took no part in running or analyzing the agents, answered a worksheet of \humanN{} verified items: five from each of \humanSetups{} setups, chosen by a seeded hash that uses neither answer letters nor model outcomes. The other setups were left out because their text does not fully determine the reference input (Appendix~\ref{sec:appendix-verification}) or because they have too few verified items. The participant worked offline with PHREEQC 3.8.6 and the frozen database, saw no answer keys and no model output, received no feedback, and could choose ``E'' when no option matched the simulation or the answer was unknown. The session took about four and a half hours.

The participant answered \humanCorrect{} of the \humanN{} items correctly (\humanAcc), chose a wrong option on \humanWrong, and chose E on \humanAbstain{} (Table~\ref{tab:human}). For each of the \humanAbstain{} the participant recorded the value obtained or the information found missing, so each followed a simulation that matched no option, a build failure in the terms of Section~\ref{sec:localize}. Of the 62 answers committed to an option, 60 were correct. The agents have no such option and must commit to one of four. This is one participant on a sample of the items, so we use it only as a reference point: an experienced user's errors on these items arise when setting up the simulation, not when reading its output.

\begin{table}[t]
\centering
\footnotesize
\begin{tabular}{lcc}
\toprule
Correct of \humanN & TOC & Inline \\
\midrule
Human expert (with PHREEQC) & \multicolumn{2}{c}{60} \\
\midrule
Claude Opus 4.6 & 64 & 64 \\
Claude Sonnet 4.6 & 68 & 67 \\
GPT-5.4 & 59 & 50 \\
GPT-5.2 & 44 & 51 \\
GPT-5.1 & 37 & 51 \\
\bottomrule
\end{tabular}
\caption{Correct answers on the \humanN{} worksheet items: the human expert, and the headline TOC and inline run of each agent on the same items.}
\label{tab:human}
\end{table}

\paragraph{\texorpdfstring{Expert review of GPT-5.4 errors.}{Expert review of GPT-5.4 errors.}}
Before the answer keys were verified and before the automatic localization of Section~\ref{sec:localize} existed, the benchmark's item author read all \expertErrN{} errors of GPT-5.4's headline TOC run on the 200-item pool and assigned each a cause (Table~\ref{tab:expert-taxonomy}); \expertErrVerN{} of these errors are on items that were later verified. In \expertPresentN{} errors (\expertPresentPct) the right value was present in the agent's output and not used, and four more were attributed to precision. The reviewer noted that this failure ``appears here where it doesn't in Opus.'' The review was made by one annotator who knew which model produced each trajectory. Only the counts of this review were retained, not the label of each error, so we cannot restrict the table to the verified items; we report it as a reading made independently of our automatic analysis that names the failure our decomposition places in closure.

\begin{table}[t]
\centering
\footnotesize
\begin{tabular}{lcc}
\toprule
Cause assigned by the reviewer & Errors & Share \\
\midrule
High-confidence wrong build & 33 & 60\% \\
Right value present, not used & 12 & 22\% \\
Precision & 4 & 7\% \\
Simulation failed & 3 & 5\% \\
Low-confidence wrong build & 2 & 4\% \\
Defective item & 1 & 2\% \\
\midrule
Total & 55 & 100\% \\
\bottomrule
\end{tabular}
\caption{Causes assigned by the domain expert to the \expertErrN{} errors of GPT-5.4's headline TOC run on the 200-item pool.}
\label{tab:expert-taxonomy}
\end{table}

%% file: main.bbl
\begin{thebibliography}{34}
\providecommand{\natexlab}[1]{#1}
\providecommand{\url}[1]{\texttt{#1}}
\expandafter\ifx\csname urlstyle\endcsname\relax
  \providecommand{\doi}[1]{doi: #1}\else
  \providecommand{\doi}{doi: \begingroup \urlstyle{rm}\Url}\fi

\bibitem[Abramovich et~al.(2025)Abramovich, Udeshi, Shao, Lieret, Xi, Milner,
  Jancheska, Yang, Jimenez, Khorrami, Krishnamurthy, Dolan-Gavitt, Shafique,
  Narasimhan, Karri, and Press]{abramovich2025enigma}
Talor Abramovich, Meet Udeshi, Minghao Shao, Kilian Lieret, Haoran Xi, Kimberly
  Milner, Sofija Jancheska, John Yang, Carlos~E. Jimenez, Farshad Khorrami,
  Prashanth Krishnamurthy, Brendan Dolan-Gavitt, Muhammad Shafique, Karthik
  Narasimhan, Ramesh Karri, and Ofir Press.
\newblock {EnIGMA}: Interactive tools substantially assist {LM} agents in
  finding security vulnerabilities.
\newblock In \emph{Proceedings of the 42nd International Conference on Machine
  Learning}, volume 267 of \emph{Proceedings of Machine Learning Research},
  pp.\  246--355. PMLR, 2025.
\newblock URL \url{https://arxiv.org/abs/2409.16165}.

\bibitem[Bogin et~al.(2024)Bogin, Yang, Gupta, Richardson, Bransom, Clark,
  Sabharwal, and Khot]{bogin2024super}
Ben Bogin, Kejuan Yang, Shashank Gupta, Kyle Richardson, Erin Bransom, Peter
  Clark, Ashish Sabharwal, and Tushar Khot.
\newblock {SUPER}: Evaluating agents on setting up and executing tasks from
  research repositories.
\newblock In \emph{Proceedings of the 2024 Conference on Empirical Methods in
  Natural Language Processing}, pp.\  12622--12645, Miami, Florida, USA, 2024.
  Association for Computational Linguistics.
\newblock \doi{10.18653/v1/2024.emnlp-main.702}.
\newblock URL \url{https://arxiv.org/abs/2409.07440}.

\bibitem[Campbell et~al.(2025)Campbell, Cox, Medina, Watterson, and
  White]{campbell2025mdcrow}
Quintina Campbell, Sam Cox, Jorge Medina, Brittany Watterson, and Andrew~D.
  White.
\newblock {MDCrow}: Automating molecular dynamics workflows with large language
  models, 2025.
\newblock URL \url{https://arxiv.org/abs/2502.09565}.

\bibitem[Cao et~al.(2026)Cao, Yin, Dhingra, and Zhou]{cao2026codingagents}
Weili Cao, Xunjian Yin, Bhuwan Dhingra, and Shuyan Zhou.
\newblock Coding agents are effective long-context processors, 2026.
\newblock URL \url{https://arxiv.org/abs/2603.20432}.

\bibitem[Chen et~al.(2024)Chen, Zhu, Zhou, and Ren]{chen2024metaopenfoam}
Yuxuan Chen, Xu~Zhu, Hua Zhou, and Zhuyin Ren.
\newblock {MetaOpenFOAM}: an {LLM}-based multi-agent framework for {CFD}, 2024.
\newblock URL \url{https://arxiv.org/abs/2407.21320}.

\bibitem[Curme \& Daugherty(2026)Curme and Daugherty]{curme2026deepagents}
Chester Curme and Mason Daugherty.
\newblock Context management for deep agents.
\newblock LangChain Blog, 28 January 2026, 2026.
\newblock URL
  \url{https://www.langchain.com/blog/context-management-for-deepagents}.

\bibitem[Druck \& Smith(2026)Druck and Smith]{druck2026rag}
Gregory Druck and Ethan Smith.
\newblock {RAG} collapse: {LLM} responses collapse when retrieved documents are
  self-authored, 2026.
\newblock URL \url{https://arxiv.org/abs/2608.22118}.

\bibitem[Engl{\"a}nder et~al.(2026)Engl{\"a}nder, Althammer, {\"U}st{\"u}n,
  Gall{\'e}, and Sherborne]{englander2026agents}
Leon Engl{\"a}nder, Sophia Althammer, Ahmet {\"U}st{\"u}n, Matthias Gall{\'e},
  and Tom Sherborne.
\newblock Agents explore but agents ignore: {LLMs} lack environmental
  curiosity, 2026.
\newblock URL \url{https://arxiv.org/abs/2604.17609}.

\bibitem[Fan et~al.(2026)Fan, Zhang, Ma, Hu, Wang, Song, Liu, Zamani, and
  Wang]{fan2026harness}
Run-Ze Fan, Zihao Zhang, Simin Ma, Yebowen Hu, Shouju Wang, Kaiqiang Song, Fei
  Liu, Hamed Zamani, and Xiaoyang Wang.
\newblock An empirical study of harness design for coding agents, 2026.
\newblock URL \url{https://arxiv.org/abs/2609.20804}.

\bibitem[Flynt(2026)]{flynt2026groundeval}
Jeffrey Flynt.
\newblock {GroundEval}: A deterministic replacement for {LLM}-as-judge in
  stateful agent evaluation, 2026.
\newblock URL \url{https://arxiv.org/abs/2606.22737}.

\bibitem[Joren et~al.(2025)Joren, Zhang, Ferng, Juan, Taly, and
  Rashtchian]{joren2025sufficient}
Hailey Joren, Jianyi Zhang, Chun-Sung Ferng, Da-Cheng Juan, Ankur Taly, and
  Cyrus Rashtchian.
\newblock Sufficient context: A new lens on retrieval augmented generation
  systems.
\newblock In \emph{International Conference on Learning Representations}, pp.\
  20310--20334, 2025.
\newblock URL \url{https://arxiv.org/abs/2411.06037}.

\bibitem[Kang et~al.(2025)Kang, Chen, Han, Inan, Wutschitz, Chen, Sim, and
  Rajmohan]{kang2025acon}
Minki Kang, Wei-Ning Chen, Dongge Han, Huseyin Inan, Lukas Wutschitz, Yanzhi
  Chen, Robert Sim, and Saravanakumar Rajmohan.
\newblock {ACON}: Optimizing context compression for long-horizon {LLM} agents,
  2025.
\newblock URL \url{https://arxiv.org/abs/2510.00615}.

\bibitem[Kate et~al.(2025)Kate, Pedapati, Basu, Rizk, Chenthamarakshan,
  Chaudhury, Agarwal, and Abdelaziz]{kate2025longfunceval}
Kiran Kate, Tejaswini Pedapati, Kinjal Basu, Yara Rizk, Vijil Chenthamarakshan,
  Subhajit Chaudhury, Mayank Agarwal, and Ibrahim Abdelaziz.
\newblock {LongFuncEval}: Measuring the effectiveness of long context models
  for function calling, 2025.
\newblock URL \url{https://arxiv.org/abs/2505.10570}.

\bibitem[Katz(2026)]{katz2026dynamic}
Jediah Katz.
\newblock Dynamic context discovery.
\newblock Cursor Blog, 6 January 2026, 2026.
\newblock URL \url{https://cursor.com/blog/dynamic-context-discovery}.

\bibitem[Lin et~al.(2026)Lin, Wang, Kwok, Guo, Nale, Fleming, and
  Cheng]{lin2026reflect}
Xiaofeng Lin, Yingxu Wang, Tung Sum~Thomas Kwok, Daniel Guo, Sahil~Arun Nale,
  Charles Fleming, and Guang Cheng.
\newblock {REFLECT}: Intervention-supported error attribution for silent
  failures in {LLM} agent traces, 2026.
\newblock URL \url{https://arxiv.org/abs/2606.09071}.

\bibitem[Lindenbauer et~al.(2025)Lindenbauer, Slinko, Felder, Bogomolov, and
  Zharov]{lindenbauer2025complexity}
Tobias Lindenbauer, Igor Slinko, Ludwig Felder, Egor Bogomolov, and Yaroslav
  Zharov.
\newblock The complexity trap: Simple observation masking is as efficient as
  {LLM} summarization for agent context management, 2025.
\newblock URL \url{https://arxiv.org/abs/2508.21433}.
\newblock NeurIPS 2025 Workshop on Deep Learning for Code in the Agentic Era.

\bibitem[Liu et~al.(2026)Liu, Mao, Zhu, and Chua]{liu2026diagnosing}
Qi~Liu, Jiaxin Mao, Fengbin Zhu, and Tat-Seng Chua.
\newblock Diagnosing search behavior and failure modes in long-horizon search
  agents, 2026.
\newblock URL \url{https://arxiv.org/abs/2608.01913}.

\bibitem[McMillan(2026)]{mcmillan2026structured}
Damon McMillan.
\newblock Structured context engineering for file-native agentic systems:
  Evaluating schema accuracy, format effectiveness, and multi-file navigation
  at scale, 2026.
\newblock URL \url{https://arxiv.org/abs/2602.05447}.

\bibitem[Mudur et~al.(2025)Mudur, Cui, Venugopalan, Raccuglia, Brenner, and
  Norgaard]{mudur2025feabench}
Nayantara Mudur, Hao Cui, Subhashini Venugopalan, Paul Raccuglia, Michael~P.
  Brenner, and Peter Norgaard.
\newblock {FEABench}: Evaluating language models on multiphysics reasoning
  ability, 2025.
\newblock URL \url{https://arxiv.org/abs/2504.06260}.
\newblock NeurIPS 2024 Workshops on Mathematical Reasoning and AI and on
  Open-World Agents.

\bibitem[Pakhomov \& Nijkamp(2026)Pakhomov and
  Nijkamp]{pakhomov2026attributing}
Egor Pakhomov and Erik Nijkamp.
\newblock Attributing agent failures to their history needs an executed
  counterfactual: Toward a reporting standard, 2026.
\newblock URL \url{https://arxiv.org/abs/2611.05650}.

\bibitem[Parkhurst \& Appelo(2013)Parkhurst and Appelo]{parkhurst2013phreeqc}
David~L. Parkhurst and C.~A.~J. Appelo.
\newblock Description of input and examples for {PHREEQC} version 3: A computer
  program for speciation, batch-reaction, one-dimensional transport, and
  inverse geochemical calculations.
\newblock Technical Report Techniques and Methods 6-A43, U.S. Geological
  Survey, 2013.

\bibitem[Petrova et~al.(2026)Petrova, Mazniak, and State]{petrova2026paginate}
Tatiana Petrova, Andrei Mazniak, and Radu State.
\newblock Agents don't paginate: First-chunk selection for {LLM} tool
  responses, 2026.
\newblock URL \url{https://arxiv.org/abs/2608.26130}.

\bibitem[Rajasekaran et~al.(2025)Rajasekaran, Dixon, Ryan, and
  Hadfield]{anthropic2025context}
Prithvi Rajasekaran, Ethan Dixon, Carly Ryan, and Jeremy Hadfield.
\newblock Effective context engineering for {AI} agents.
\newblock Anthropic Engineering Blog, 29 September 2025, September 2025.
\newblock URL
  \url{https://www.anthropic.com/engineering/effective-context-engineering-for-ai-agents}.

\bibitem[Roh \& Han(2026)Roh and Han]{roh2026before}
Daeyoung Roh and Donghee Han.
\newblock Before reasoning can fail: Pre-evidence procedural failures in
  agentic {RAG}, 2026.
\newblock URL \url{https://arxiv.org/abs/2608.02011}.

\bibitem[Ru et~al.(2024)Ru, Qiu, Hu, Zhang, Shi, Chang, Jiayang, Wang, Sun, Li,
  Zhang, Wang, Jiang, He, Wang, Liu, Zhang, and Zhang]{ru2024ragchecker}
Dongyu Ru, Lin Qiu, Xiangkun Hu, Tianhang Zhang, Peng Shi, Shuaichen Chang,
  Cheng Jiayang, Cunxiang Wang, Shichao Sun, Huanyu Li, Zizhao Zhang, Binjie
  Wang, Jiarong Jiang, Tong He, Zhiguo Wang, Pengfei Liu, Yue Zhang, and Zheng
  Zhang.
\newblock {RAGChecker}: A fine-grained framework for diagnosing
  retrieval-augmented generation.
\newblock In \emph{Advances in Neural Information Processing Systems (Datasets
  and Benchmarks Track)}, volume~37, pp.\  21999--22027, 2024.
\newblock \doi{10.52202/079017-0692}.
\newblock URL \url{https://arxiv.org/abs/2408.08067}.

\bibitem[Sen et~al.(2026)Sen, Kasturi, Lumer, Gulati, and Subbiah]{sen2026grep}
Sahil Sen, Akhil Kasturi, Elias Lumer, Anmol Gulati, and Vamse~Kumar Subbiah.
\newblock Is grep all you need? {H}ow agent harnesses reshape agentic search,
  2026.
\newblock URL \url{https://arxiv.org/abs/2605.15184}.

\bibitem[Tan et~al.(2024)Tan, Sun, Yang, Wang, Cao, and Cheng]{tan2024blinded}
Hexiang Tan, Fei Sun, Wanli Yang, Yuanzhuo Wang, Qi~Cao, and Xueqi Cheng.
\newblock Blinded by generated contexts: How language models merge generated
  and retrieved contexts when knowledge conflicts?
\newblock In \emph{Proceedings of the 62nd Annual Meeting of the Association
  for Computational Linguistics (Volume 1: Long Papers)}, pp.\  6207--6227,
  Bangkok, Thailand, 2024. Association for Computational Linguistics.
\newblock \doi{10.18653/v1/2024.acl-long.337}.
\newblock URL \url{https://arxiv.org/abs/2401.11911}.

\bibitem[Xie et~al.(2026)Xie, Zhou, Fan, Ge, Han, and Liu]{xie2026bayes}
Yi~Xie, Zhanke Zhou, Yi~Fan, Yong Ge, Bo~Han, and Bo~Liu.
\newblock Bayes-sufficient compression is not enough: How does communication
  help multi-agent systems?
\newblock In \emph{Advances in Neural Information Processing Systems
  (NeurIPS)}, 2026.
\newblock URL \url{https://arxiv.org/abs/2610.03769}.
\newblock Preprint, arXiv:2610.03769.

\bibitem[Yang et~al.(2025)Yang, Hu, Kang, Lin, and
  Zhang]{yang2025numbercookbook}
Haotong Yang, Yi~Hu, Shijia Kang, Zhouchen Lin, and Muhan Zhang.
\newblock {Number Cookbook}: Number understanding of language models and how to
  improve it.
\newblock In \emph{International Conference on Learning Representations}, pp.\
  3946--3984, 2025.
\newblock URL \url{https://arxiv.org/abs/2411.03766}.

\bibitem[Yang et~al.(2024)Yang, Jimenez, Wettig, Lieret, Yao, Narasimhan, and
  Press]{yang2024sweagent}
John Yang, Carlos~E. Jimenez, Alexander Wettig, Kilian Lieret, Shunyu Yao,
  Karthik Narasimhan, and Ofir Press.
\newblock {SWE}-agent: Agent-computer interfaces enable automated software
  engineering.
\newblock In \emph{Advances in Neural Information Processing Systems}, 2024.

\bibitem[Yue et~al.(2026)Yue, Somasekharan, Zhang, Cao, Chen, Di, and
  Pan]{yue2026foamagent}
Ling Yue, Nithin Somasekharan, Tingwen Zhang, Yadi Cao, Zhangze Chen, Shimin
  Di, and Shaowu Pan.
\newblock {Foam-Agent}: A large language model-based multi-agent framework for
  automating computational fluid dynamics workflows.
\newblock \emph{Computer Methods in Applied Mechanics and Engineering},
  461:\penalty0 119271, 2026.
\newblock URL \url{https://arxiv.org/abs/2505.04997}.

\bibitem[Zhang et~al.(2026{\natexlab{a}})Zhang, Wang, Chen, Zhou, Wang, and
  Yan]{zhang2026agentracer}
Guibin Zhang, Junhao Wang, Junjie Chen, Wangchunshu Zhou, Kun Wang, and
  Shuicheng Yan.
\newblock {AgenTracer}: Who is inducing failure in the {LLM} agentic systems?
\newblock In \emph{International Conference on Learning Representations}, pp.\
  11377--11399, 2026{\natexlab{a}}.
\newblock URL \url{https://arxiv.org/abs/2509.03312}.

\bibitem[Zhang et~al.(2026{\natexlab{b}})Zhang, Chundur, Qomi, and
  Raissi]{zhang2026phreeqcmcq}
Ke~Zhang, Sahchit Chundur, Mohammad~Javad Qomi, and Maziar Raissi.
\newblock {PHREEQC-MCQ-200}: A diagnostic benchmark for tool-augmented
  scientific simulator agents, 2026{\natexlab{b}}.
\newblock URL \url{https://arxiv.org/abs/2607.00436}.

\bibitem[Zhang et~al.(2025)Zhang, Yin, Zhang, Liu, Han, Zhang, Li, Wang, Wang,
  Chen, and Wu]{zhang2025whoandwhen}
Shaokun Zhang, Ming Yin, Jieyu Zhang, Jiale Liu, Zhiguang Han, Jingyang Zhang,
  Beibin Li, Chi Wang, Huazheng Wang, Yiran Chen, and Qingyun Wu.
\newblock Which agent causes task failures and when? {On} automated failure
  attribution of {LLM} multi-agent systems.
\newblock In \emph{Proceedings of the 42nd International Conference on Machine
  Learning}, volume 267 of \emph{Proceedings of Machine Learning Research},
  pp.\  76583--76599. PMLR, 2025.
\newblock URL \url{https://arxiv.org/abs/2505.00212}.

\end{thebibliography}
